%% file: main.tex
\documentclass{article}

\usepackage{microtype}
\usepackage{graphicx}
\usepackage{subfigure}
\usepackage{booktabs} % for professional tables

\usepackage{hyperref}

\usepackage[accepted]{icml2025}
\makeatletter
\renewcommand{\Notice@String}{} % removes the “Proceedings of ...” line
\makeatother
\input{math_commands}

\usepackage{amsmath}
\usepackage{amssymb}
\usepackage{mathtools}
\usepackage{amsthm}

\usepackage[capitalize,noabbrev]{cleveref}

\usepackage{multirow}
\usepackage{capt-of}
\usepackage{subcaption}
\usepackage{xspace}

\usepackage{tikz}
\newcommand*\circled[1]{\tikz[baseline=(char.base)]{
          \node[shape=circle,draw,inner sep=1pt] (char) {#1};}}
\newcommand{\ours}{ScaleBITS\xspace} %\xspace
\usepackage{colortbl}
\newcommand{\myrowcolour}{\rowcolor[gray]{0.925}}

\theoremstyle{plain}

\theoremstyle{definition}

\theoremstyle{remark}

\usepackage[textsize=tiny]{todonotes}

\icmltitlerunning{Scalable Bitwidth Search for Hardware-Aligned Mixed-Precision LLMs}

\begin{document}

\twocolumn[
\icmltitle{ScaleBITS: Scalable Bitwidth Search for Hardware-Aligned \\ Mixed-Precision LLMs}




\begin{icmlauthorlist}
\icmlauthor{Xinlin Li}{msl,ucla}
\icmlauthor{Timothy Chou}{msl}
\icmlauthor{Josh Fromm}{msl}
\icmlauthor{Zichang Liu}{msl}
\icmlauthor{Yunjie Pan}{meta}
\icmlauthor{Christina Fragouli}{ucla}
\end{icmlauthorlist}

\icmlaffiliation{ucla}{University of California, Los Angeles}
\icmlaffiliation{msl}{Meta Superintelligence Labs, Meta Platforms, Inc.}
\icmlaffiliation{meta}{Meta Platforms, Inc.}

\icmlcorrespondingauthor{Xinlin Li}{xinlinli@ucla.edu}

\icmlkeywords{Machine Learning, ICML}

\vskip 0.3in
]



\printAffiliationsAndNotice{}

\begin{abstract}
Post-training weight quantization is crucial for reducing the memory and inference cost of large language models (LLMs), yet pushing the average precision below 4 bits remains challenging due to highly non-uniform weight sensitivity and the lack of principled precision allocation. Existing solutions use irregular fine-grained mixed-precision with high runtime overhead or rely on heuristics or highly constrained precision allocation strategies. In this work, we propose \ours, a mixed-precision quantization framework that enables automated, fine-grained bitwidth allocation under a memory budget while preserving hardware efficiency. Guided by a new sensitivity analysis, we introduce a hardware-aligned, block-wise weight partitioning scheme, powered by bi-directional channel reordering. We formulate global bitwidth allocation as a constrained optimization problem and develop a scalable approximation to the greedy algorithm, enabling end-to-end principled allocation. Experiments show that \ours significantly improves over uniform-precision quantization (up to +36\%) and outperforms state-of-the-art sensitivity-aware baselines (up to +13\%) in ultra-low-bit regime, without adding runtime overhead.
\end{abstract}

\section{Introduction}\label{sec:intro}
\begin{figure}[t]
    \centering
    \includegraphics[width=0.7\linewidth]{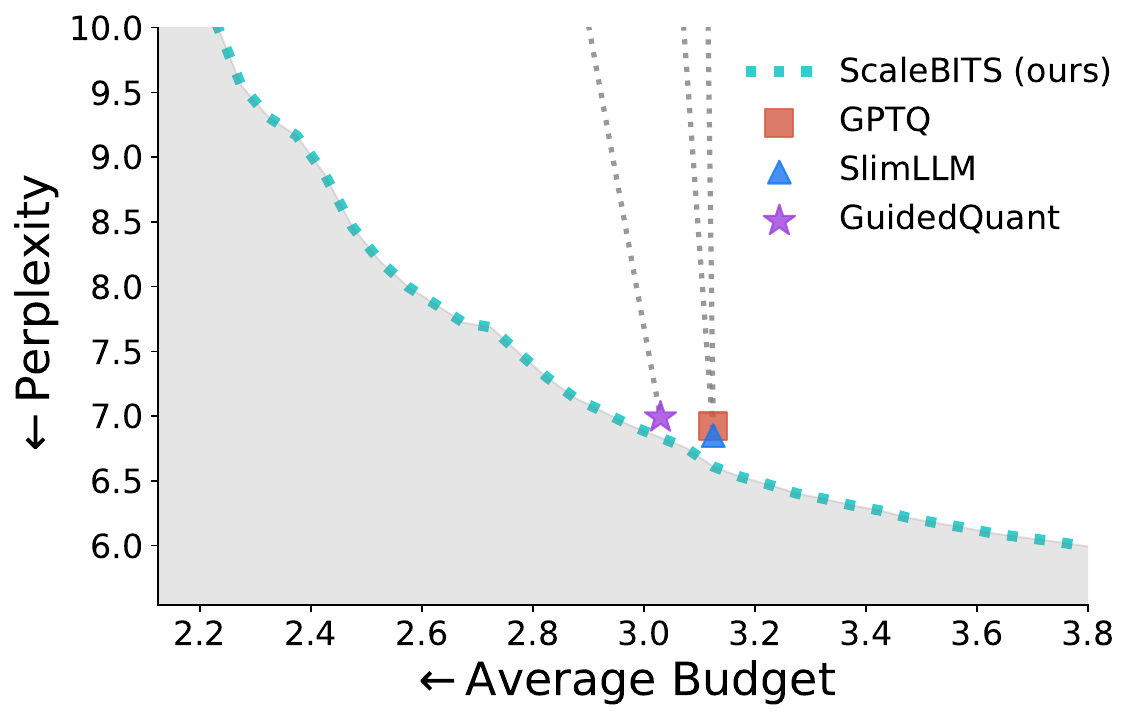}
    \vspace{-.1in}
    \caption{\ours yields a smooth accuracy–compression trade-off beyond the budgets supported by existing methods.} %Llama3-8B model in ultra low-bit regime.}
    \label{fig:accuracy}
\end{figure}

The rapid scaling of large language models (LLMs) has made post-training weight-only quantization an essential technique for reducing memory footprint and inference cost. While standard 4-bit formats (e.g., INT4 and FP4) are often sufficient to preserve model quality, pushing weight quantization below 4 bits remains challenging, as accuracy degrades rapidly. Recent solutions increasingly rely on unstructured mixed-precision or complex vector quantization schemes, which improve quantization quality at the cost of additional runtime overhead. In this work, we investigate whether more principled mixed-precision allocation can better exploit simple ultra-low-bit representations, without sacrificing runtime efficiency.

A key difficulty in the ultra-low-bit regime is the highly non-uniform sensitivity of model weights. While most weights tolerate aggressive quantization, errors on a small subset of critical weights can severely degrade model behavior \cite{super-weight, sunmassive}. Existing approaches address this through either uniform or mixed-precision. 

Uniform-precision methods maintain a fixed bitwidth while reducing quantization error on sensitive weights through equivalent transformations \cite{lin2024awq}, error compensation \cite{frantar2023gptq}, adaptive rounding \cite{chee2023quip}, or improved quantization grids \cite{kim2023squeezellm,kim2025guidedquant}. Although effective at moderate precision, these methods fundamentally saturate in the ultra-low-bit regime, where the limited number of representation levels makes it difficult to balance quantization error across heterogeneous weights. Moreover, uniform-precision methods are restricted to a small set of discrete operating points (e.g., 2, 3, or 4 bits), which severely limits the granularity of achievable compression–accuracy trade-offs, especially when the gap between adjacent bitwidths is large.

Mixed-precision quantization offers a more direct way to address non-uniform weight sensitivity by assigning higher precision to more important components. However, designing a practical mixed-precision solution for LLMs remains challenging due to two requirements: (i) hardware-friendly granularity, and (ii) scalable precision allocation strategy. At one extreme, coarse, layer-wise precision assignment is computationally efficient and has been explored extensively in earlier work \cite{wang2019haq, dong2020hawq, chen2021towards}. However, it is inadequate for LLMs, where each layer contains millions of parameters and the sensitive weights are sparsely distributed. At the other extreme, recent fine-grained schemes \cite{dettmers2023spqr,kim2023squeezellm,yuan2024pbllm} preserve a small fraction of sensitive weight elements in full precision, achieving strong accuracy but introducing irregular sparsity, index overhead and poor hardware utilization. Intermediate granularities, such as channel-grouped partitions \cite{lee2024owq,huang2024slim}, have been explored, but their designs inevitably trade off adaptivity for runtime regularity.

Beyond partition granularity, the effectiveness of mixed-precision quantization critically depends on how precision levels are determined for each weight group. This allocation problem is challenging due to the combinatorial search space, which grows rapidly with model size, partition granularity, and the number of candidate precisions. Classic mixed-precision methods \cite{cai2020rethinking,wang2019haq,yuan2020evoq}—based on differentiable search, reinforcement learning, or evolutionary algorithms—are computationally prohibitive at LLM scale and typically require retraining or extensive evaluation. As a result, most LLM quantization pipelines rely on sensitivity-guided heuristics, manually specified precision ratios, or severely restricted search spaces \cite{dettmers2023spqr,kim2023squeezellm,hu2025mlwq,huang2024slim}.

In this work, we revisit mixed-precision quantization from a unified optimization perspective, framing it as a discrete resource allocation problem. Under mild assumptions, this formulation admits a classic greedy solution with provable approximation guarantees. However, applying such a greedy strategy directly at the scale and granularity of modern LLMs is impractical, as it would require an excessive number of sequential evaluations to obtain the marginal gains of precision adjustments. Our key insight is that greedy decisions are driven primarily by the relative ordering of these marginal gains, which can be efficiently approximated using a lightweight sensitivity estimate computed around a progressively quantized model. Our sensitivity analysis further reveals a bi-directional structural pattern: sensitive weights are concentrated in a small subset of both output (row) and input (column) channels, forming localized high-sensitivity regions within weight matrices. This structure indicates that suitably reordered, block-wise partitions can preserve high expressivity.

Building on these insights, we introduce \ours, a scalable mixed-precision quantization framework for LLMs, which achieves a continuous Pareto frontier, reaching operating points that are inaccessible to existing uniform or locally constrained mixed-precision schemes (as shown in Figure~\ref{fig:accuracy}). \ours first applies a bi-directional channel reordering strategy to cluster sensitive weights into contiguous regions within each matrix. It then employs hardware-aligned, block-wise partitions that map efficiently to optimized matrix-multiplication kernels, enabling uniform execution within each block. Finally, we propose a scalable approximation to the classic greedy algorithm, allowing automated precision allocation under an arbitrary global memory budget. Our main contributions are:
\begin{itemize}
    \item A progressive-quantization sensitivity analysis that exposes both the dynamic behavior of sensitivities and a previously underexplored bi-directional structure across weight channels.
    \item A hardware-aligned, block-wise weight partitioning strategy via bi-directional channel reordering, jointly optimizing expressiveness and runtime efficiency.
    \item A scalable, automated precision search algorithm that enables fine-grained mixed-precision allocation for large language models without retraining.
    \item Extensive experiments demonstrating that our framework allows even a simple round-to-nearest uniform quantizer to outperform state-of-the-art sensitivity-aware and mixed-precision methods in the ultra-low-bit regime, with no additional inference overhead.
\end{itemize}

\section{Problem Formulation}\label{sec:problem}
\paragraph{Notation.}
This paper focuses on weight-only quantization. Given a pre-trained model, we use $\vw$ to denote the full set of model weights and $\Pi_\vw = \{\vw_i\}_{i=1}^N$ to denote a partition of $\vw$, where $\vw_i \in \mathbb{R}^{M_i}$ is a vectorization of all weights in component $i$, $N$ is the total number of components and $M_i$ is the size of each. For example, $\vw_i$ may correspond to weights in a linear layer or a channel group (i.e., row/column in a weight matrix). The finest partition corresponds to $M_i=1$, where each component reduces to a scalar weight element $w_i \in \mathbb{R}$. When the partition $\Pi_\vw$ is clear from context, we use $\vw$ in expressions for simplicity.

We assume that all weights in the same component share the same precision (e.g., 2 bits per weight). We use $\vb = (b_1,\ldots, b_N)\in 
\gB^N$ to denote the precision assigned to each component, where  $\gB$ is the set of candidate precisions. In the most general case $\gB = \mathbb{Z}_{\ge 0}^N$. We note that if $b_i=0$, the component $\vw_i$ is effectively pruned. We use $Q(\cdot, \cdot)$ to denote a quantization scheme (or quantizer), and $Q(\vw,\vb)$ is the quantized model under precision $\vb$.

Let $\ell(\vx;\vw)$ be the per-example loss. We denote the population loss as $L(\vw) = \mathbb{E}_{\vx}[\ell(\vx;\vw)]$ and use $L_{\gD}(\vw)$ for the empirical loss evaluated on a calibration dataset $\gD$. 

\subsection{Problem Formulation}
Mixed-precision quantization involves two core problems: (i) designing the weight partition $\Pi_\vw$ and (ii) optimizing the bit allocation $\vb$. 

The weight partition $\Pi_{\vw}$ determines the mixed-precision granularity and must balance quantization quality and runtime efficiency. From a quality perspective, an ideal partition groups weights with similar sensitivity, reducing sensitivity variation within each component and enabling appropriate bitwidth sharing.
Finer-grained partitions are more expressive, as they allow precision to adapt to local sensitivity differences; in the extreme, per-element precision assignment offers maximum flexibility.
However, unstructured partitions typically incur additional overhead in metadata storage, control flow, and memory access, preventing efficient runtime execution. The core challenge is therefore to design $\Pi_{\vw}$ that maximizes sensitivity clustering while maintaining compatibility with efficient hardware execution.

Given a weight partition $\Pi_\vw$ and a total memory budget $B$ (bits per weight), finding an optimal bit allocation $\vb^*$ can be formulated as the following integer optimization problem
\begin{equation}\label{eq:formulation}
    \begin{aligned}\min_\vb \quad & L(Q(\vw,\vb))\\\text{s.t.} \quad &\frac{1}{N}\sum_i^N b_i \le B, \\ & \vb \in \mathbb{Z}_{\ge 0} ^N. \end{aligned}
\end{equation}
Directly solving this problem is challenging due to the non-convex objective function and the discrete yet vast search space. Under some natural assumptions (see Appendix~\ref{app:assumption}), it becomes optimizing a monotone submodular function over the integer lattice with a knapsack constraint, which is a well-studied NP-hard problem \cite{nemhauser1978best}.
%This class of problems is well-studied in combinatorial optimization and is known to be NP-hard \cite{nemhauser1978best}. 
A classic greedy algorithm provides a simple solution with $(1-1/e)$ approximation guarantee to the optimal $\vb^*$ \cite{nemhauser1978analysis} and this bound is tight, meaning no polynomial-time algorithm can achieve a better worst-case guarantee \cite{feige1998threshold}. For completeness, we restate the classic greedy search in Algorithm~\ref{alg:cgreedy} for mixed-precision allocation, which has been shown effective in the layer precision search for small CNNs \cite{chen2021towards}.

Despite its simple concept and strong guarantee, classic greedy search is prohibitively expensive at scale. It starts from the minimal precision and iteratively increases the precision of a single weight component $\vw_i$ that produces the largest marginal decrease in loss. To reach the target budget $B$, the number of required evaluations (each involving forward passes over a calibration dataset) scales as $O(N^2)$. In practice, $N$ becomes extremely large due to the combined effect of larger model size and fine partition. For example, partitioning the Gemma2-9B model into blocks of size $M = 64 \times 64$ results in approximately $N \approx \frac{9\times 10^9}{4096} \approx 2.2 \times 10^6$. This motivates the need for scalable approximations that preserve the greedy decision structure without explicitly evaluating all marginal losses.

\begin{table}[t]
    \centering
     \resizebox{0.48\textwidth}{!}{
    \begin{tabular}{c|c|l}
    \toprule
         \textbf{Method} & \textbf{MP Granularity} &  \textbf{Sensitivity Metric} \\
    \midrule
    LLM-MQ & layer & \circled{1} $\begin{aligned}
         |g^{(w_i)}|\,|\Delta w_i| \end{aligned}$ \\
    \midrule
    TACQ & element & $ \circled{2} \begin{aligned}\;
        |g^{(w_i)}|\,|\Delta w_i|\,|w_i|\end{aligned}$\\
    \midrule
         SqueezeLLM & element & \multirow{1}{*}{\circled{3} $\begin{aligned}\; 
         \gF_{ii}\, \Delta w_i^{\,2}
         %(\frac{\partial L(\vw)}{\partial w_i})^2 \Delta w_i^2
         \end{aligned}$} \\
         % MixLLM & channel & \\
    \midrule
         SpQR, PB-LLM & element & \multirow{2}{*}{\circled{4} $\begin{aligned}\; \frac{(\Delta w_i)^2}{[\mX\mX^\top]^{-1}_{ii}} \end{aligned}$}\\
         OWQ, SlimLLM & channel (group) & \\
    \bottomrule
    \end{tabular}
    }
    \vspace{-.05in}
    \captionof{table}{Comparison of commonly used sensitivity metrics for a single weight $w_i \in \mathbb{R}$. Here %$g_i=\mathbb{E}[\frac{\partial l(\vw)}{\partial w_i}]$, 
    $\gF_{ii} \approx \mathbb{E}[(\frac{\partial\ell(\vw)}{\partial w_i})^2]$, and $\mX$ is the batched input activation to the linear layer. For a large weight component $\vw_i \in \mathbb{R}^{M_i}$ with $M_i > 1$, sensitivity is typically aggregated by summation or averaging.} %averaging over all weights.}
    \label{tab:sensitivity}
\end{table}

\begin{figure*}[h]
    \centering    \includegraphics[width=0.65\linewidth]{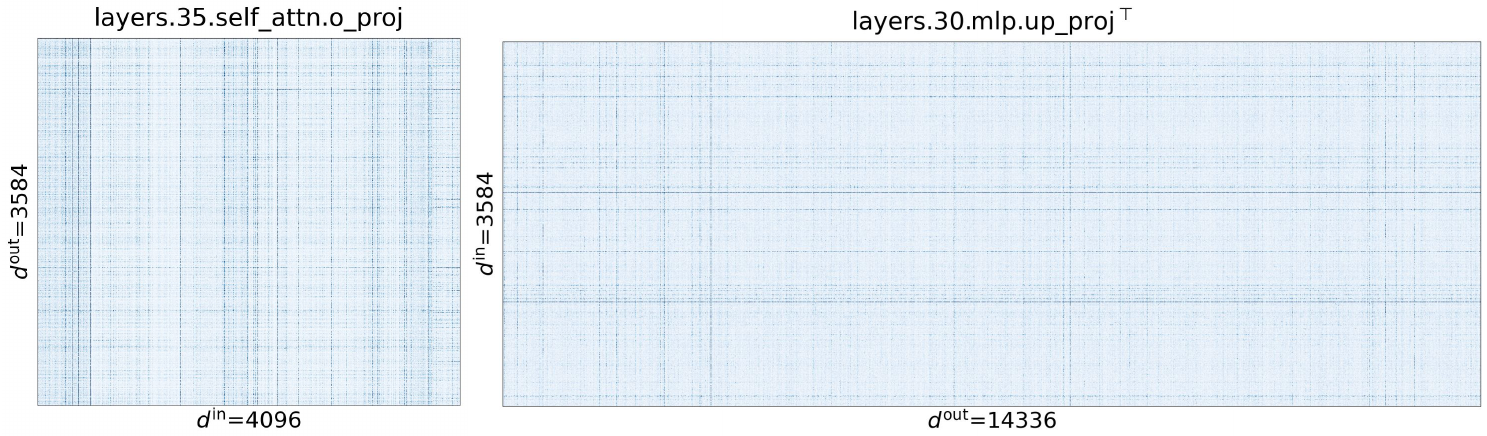}
    \vspace{-.1in}
    \caption{Weight sensitivity distribution in Gemma2-9B model.}
    \label{fig:spatial}
    \vspace{-.15in}
\end{figure*}

\section{Revisit Weight Sensitivity}\label{sec:sensitivity}
To scale the greedy mixed-precision optimization in Section~\ref{sec:problem}, we require an efficient proxy for the marginal loss changes induced by adjusting the precision. In this section, we revisit weight sensitivity from this optimization perspective, and study how it can be estimated reliably under progressive quantization. Beyond optimization, the spatial structure of the sensitivity distribution also inspires our design of hardware-aligned weight partitions.

\subsection{Sensitivity Estimation}
The \textit{sensitivity/saliency} of a weight component $\vw_i$ measures the degradation of model quality (final loss) induced by modifying that component, for example, through quantization or pruning. At fine granularity, enumerating all component-wise changes and obtaining the exact loss difference is computationally infeasible. As a result, the Taylor series is commonly used to approximate the impact of small perturbations $\Delta \vw_i$. 

Most existing methods expand the loss around the full-precision model (with weights $\vw$):
\begin{equation}\label{eq:taylor}
\begin{aligned}
& |L(\vw + \Delta \vw_i) - L(\vw)| \\
\approx~ & |{\vg^{(\vw_i)}}^\top\Delta \vw_i + \frac{1}{2}\Delta \vw_i^\top H^{(\vw_i)} \Delta \vw_i|,
\end{aligned}
\end{equation}
where $\vg^{(\vw_i)}=\nabla_{\vw_i}L(\vw)$ and $H^{(\vw_i)}=\nabla_{\vw_i}^2L(\vw)$ denote the gradient and Hessian with respect to $\vw_i$. Table~\ref{tab:sensitivity} summarizes representative sensitivity metrics derived from this approximation. Some approaches retain only the first-order term \cite{li2023llm,xiao2025task}, while others assume the pre-trained model is near a local minimum (i.e., $\vg^{(\vw_i)} \approx \vzero$) and rely solely on the second-order information \cite{kim2023squeezellm,dettmers2023spqr,yuan2024pbllm,lee2024owq,huang2024slim}. Because computing the full Hessian $H^{(\vw_i)}$ is intractable for LLMs, the curvature is further approximated using local activation statistics or the diagonal Fisher information matrix. These approaches implicitly assume that derivatives evaluated at the full-precision model remain informative after quantization.

However, weight sensitivity is inherently dynamic: the impact of perturbing one component $\vw_i$ depends on the precision states of the entire model. This dependency becomes especially important in the ultra-low-bit regime. When most weights are quantized aggressively, the resulting model can be far from the full-precision one, and the loss landscape may change substantially. As a result, sensitivity measured at the full-precision model may no longer reflect the true effect of additional perturbation. 

To address this issue, we propose to estimate sensitivity with respect to a quantized reference point, which is naturally available when the model is progressively quantized (as in Algorithm~\ref{alg:cgreedy} and \ref{alg:sgreedy}). Specifically, instead of using the full-precision model, we expand the loss around the quantized model (with weights $\vw^Q$) and define the sensitivity of a component $\vw_i$ as
\begin{equation}\label{eq:sensitivity}
s_i :=~|L(\vw^Q + \Delta \vw_i) - L(\vw^Q)| \approx~|{\vg^{(\vw^Q_i)}}^\top\Delta \vw_i| %+ \frac{1}{2}\Delta \vw_i^\top H_{w^Q} \Delta \vw_i,
\end{equation}
where $\vg^{(\vw^Q_i)} = \nabla_{\vw^Q_i}L(\vw^Q)$. This choice offers two advantages. First, since $\vw^Q$ is closer to the current quantized model, the local derivatives better capture the true loss behavior under quantization. Second, unlike the full-precision model, $\vw^Q$ are not at a local optimum with respect to the model loss; as a result, the first-order term dominates and can be estimated reliably without resorting to expensive second-order approximation. 

\begin{figure}[h]
    \centering
    \includegraphics[width=0.8\linewidth]{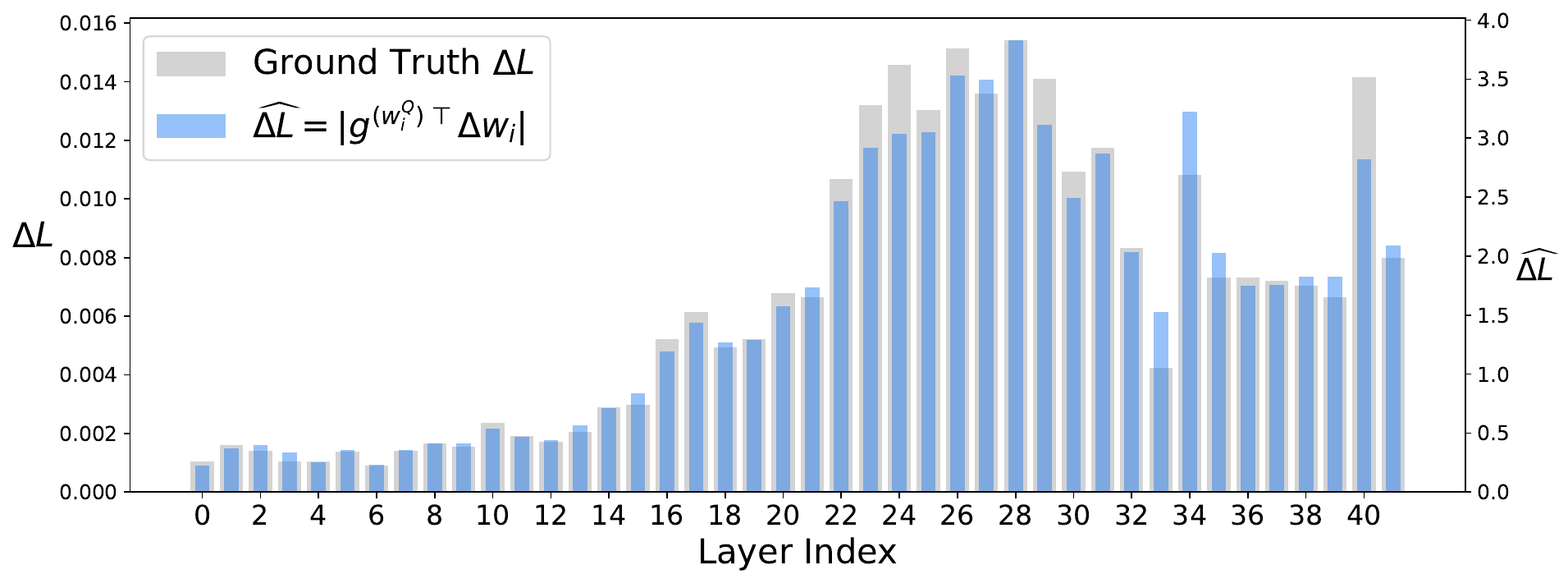} \\
    \includegraphics[width=0.8\linewidth]{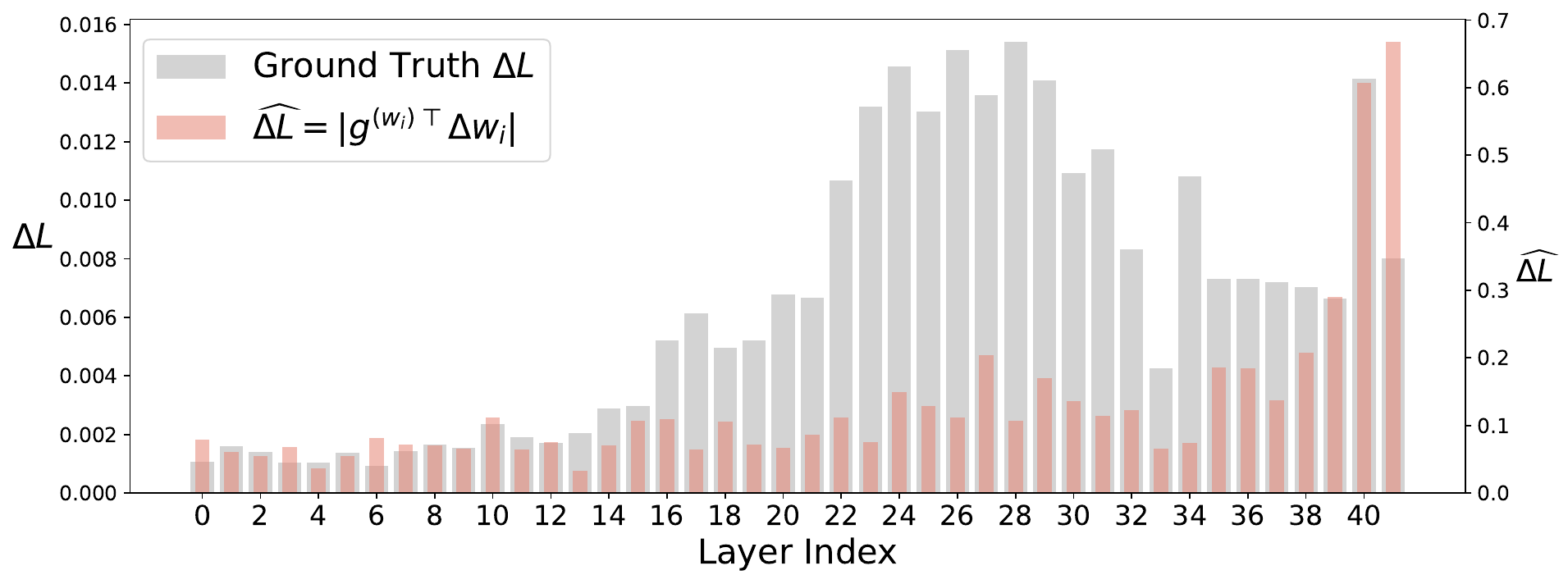}
    \vspace{-.1in}
    \caption{Estimated layer sensitivity using first-order Taylor expansion around a quantized model (top) and the full-precision model (bottom).}
    \label{fig:layer_sensitivity}
    \vspace{-.1in}
\end{figure}

Figure~\ref{fig:layer_sensitivity} validates this design by comparing layer-wise sensitivities (where $\vw_i$ corresponds to a decoder layer) obtained from gradients evaluated at the quantized model (Eq.~\ref{eq:sensitivity}) versus the full-precision one (\circled{1} in Table~\ref{tab:sensitivity}). The ground-truth loss changes are computed by restoring a single layer to higher precision while keeping the rest quantized. We found that the estimate using the quantized model effectively preserves the correct ordering of layer sensitivities, although it tends to overestimate the magnitude. In contrast, the full-precision estimate fails to produce meaningful rankings. Comparison with other metrics is provided in Appendix~\ref{app:sensitivity}.

\begin{figure*}[t]
    \centering
    \includegraphics[width=0.7\linewidth]{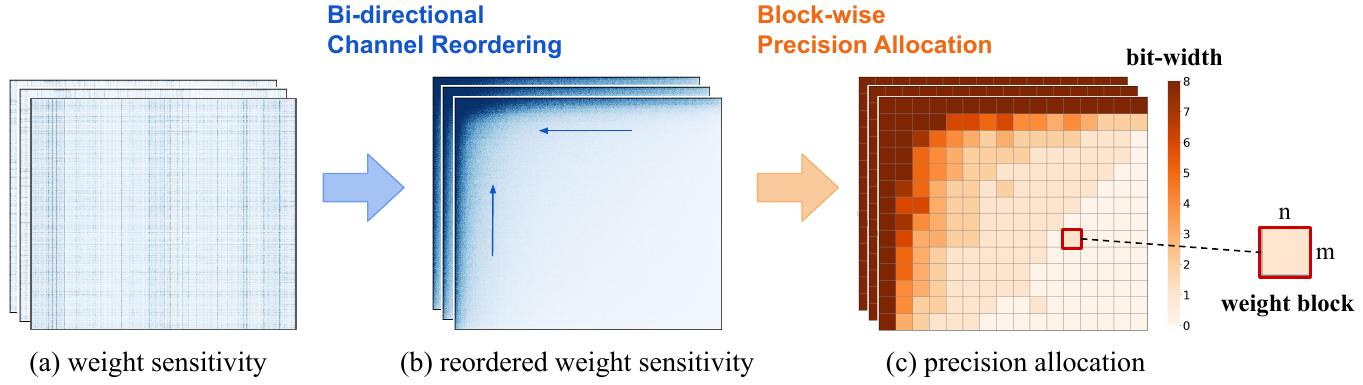}
    \vspace{-.12in}
    \caption{Overview of the proposed mixed-precision framework. 
    (a) Sensitivity distribution of the original weight matrix.
    (b) Sensitivity distribution after bi-directional reordering of input and output channels based on sensitivity.
    (c) Resulting block-wise precision allocation over the reordered weight matrix.}
    \label{fig:overview}
    \vspace{-.15in}
\end{figure*}

\subsection{Spatial Distribution of Sensitive Weights}\label{sec:spatial}
Having established the reliability of our sensitivity estimate at the layer level, we focus on individual weight elements ($w_i \in \mathbb{R}$) and study the spatial distribution of sensitive weights within each matrix. Using our sensitivity metric (Eq.~\ref{eq:sensitivity}), we visualize the spatial distribution of weight sensitivities in each weight matrix. Figure~\ref{fig:spatial} shows examples from Gemma2-9B model, where sensitive weights concentrate in a few columns and rows (see more examples in Appendix~\ref{app:spatial}). This pattern can be explained by expanding the gradient term in Eq.~\ref{eq:sensitivity}. Given a linear layer with output $\vy=\mW^Q\vx$, where $\mW^Q$ is the quantized weights in matrix form, the gradient of each entry $W_{ij}^Q$ is calculated as
\begin{equation}
    g^{(\mW^Q)}_{ij} := \frac{\partial L}{\partial W_{ij}^Q} = g_i^{(\vy)}x_j,
\end{equation}
where $g_i^{(\vy)}$ is the output gradient of row $i$ and $x_j$ is the input activation of column $j$. Plugging this into the sensitivity metric, we have 
\begin{equation}
    s_{ij} = |g^{(\mW^Q)}_{ij}\Delta W_{ij}| = |g_i^{(\vy)}|\;|x_j|\;|\Delta W_{ij}|,
\end{equation}
which consists of a row-wise factor $|g_i^{(\vy)}|$, a column-wise factor $|x_j|$ and a local weight distortion $|\Delta W_{ij}|$.

This two-dimensional pattern inspires our partition scheme presented in Section~\ref{sec:partition}. Note that prior work, such as SlimLLM \cite{huang2024slim} reports only column-wise clustering, likely because their sensitivity metric \circled{4} overlooks the inter-layer dependency captured in output gradients $\vg^{(\vy)}$. 

\section{Methodology}\label{sec:method}
In this section, we introduce a hardware-aligned mixed-precision quantization framework that builds on the sensitivity analysis in Section~\ref{sec:sensitivity} and addresses the two problems stated in Section~\ref{sec:problem}. As shown in Figure~\ref{fig:overview}, we first reorder the channels of weight matrices by sensitivity, clustering the most sensitive weight toward the top-left corner. We then partition the reordered weight matrices into small blocks that match efficient matrix-multiplication kernels. Finally, precision allocation among these blocks is optimized through a scalable automated search algorithm. 

\subsection{Hardware-aligned Weight Partition}\label{sec:partition}
\textbf{Block-wise Weight Partition.} In the deployment of weight-only quantization, weights are dequantized on the fly, and matrix multiplication is performed in high precision. For efficient inference, it is desirable to fuse dequantization with matrix multiplication so that dequantized weights are consumed immediately, avoiding intermediate storage and additional memory traffic. Modern accelerators achieve high throughput by decomposing large matrix multiplications into fixed-size tiles (e.g., $64\times64$ blocks) that are executed in parallel, with each tile following a uniform and predetermined sequence of operations. Specialized units such as Tensor Cores are designed to operate most efficiently under this tiled execution structure.

Motivated by this execution behavior, we adopt weight blocks as the basic unit for precision assignment.  In particular, we choose the weight partition $\Pi_{\vw} = \{\vw_i\}_{i=1}^N$ such that $\vw_i=\text{vec}(\mW_i)$ with $\mW_i \in \mathbb{R}^{m\times n}$ being a submatrix of some weight matrix in the model.  Each block is assigned a single precision level, ensuring that dequantization and subsequent matrix multiplication within a tile follow a uniform execution pattern. In practice, the block size can be chosen to match the underlying matrix multiplication kernel and may be coarsened by grouping multiple tiles together, but it cannot be finer than the kernel’s computation granularity.

\textbf{Bi-directional Channel Reordering.} As observed in Section~\ref{sec:sensitivity}, sensitive weights tend to concentrate along a small subset of input and output channels distributed across the weight matrix. Under a block-wise layout, such a structure may be diluted unless channels with similar sensitivity are grouped together. To better capture this structure under a block-wise layout, we propose to reorder both rows (output channels) and columns (input channels) of each weight matrix based on sensitivity. Specifically, for each channel, we aggregate element-wise sensitivity using the $\ell_1$ norm, which emphasizes the presence of highly sensitive weight elements within the channel rather than canceling them out. After reordering, sensitive weights concentrate toward the top-left region of the matrix, where higher precision is preferentially allocated, as shown in Figure~\ref{fig:overview}~(b).
 
This reordering is a one-time preprocessing step performed before quantization and introduces no inference-time overhead.
To preserve functional equivalence, channel reordering must be applied consistently across connected layers. For example, the output channel order of one linear layer must match the input channel order of the next. Empirically, channels linked by these constraints tend to have similar sensitivity statistics, making joint reordering effective. Further details are provided in Appendix~\ref{app:reorder}.

Although (input) channel reordering \cite{li2025icquant, frantar2023gptq} and block-wise quantization \cite{jang2025blockdialect, hooper2025fgmp} have been explored independently for other purposes, our key insight is that reordering in \textit{both directions} is essential for making coarse, hardware-aligned weight partitions expressive enough to support mixed-precision allocation in large models.

\subsection{Scalable Precision Search}
Given a fixed weight partition $\Pi_\vw$, the remaining problem is to allocate per-block precisions $\vb \in \mathbb{Z}_{\ge 0}^N$ under a global bit budget $B$. As discussed in Section~\ref{sec:problem}, this problem admits a greedy solution with theoretical guarantees. However, directly applying the classic greedy search (Algorithm~\ref{alg:cgreedy}) is computationally infeasible for such fine-grained, block-wise precision allocation in large models. %, as it requires sequentially evaluating marginal gains for millions of blocks. 
In this section, we construct a scalable approximation that preserves the core structure of greedy search while dramatically reducing its computational cost. The resulting scalable search procedure is summarized in Algorithm~\ref{alg:sgreedy}.
 
At each iteration, the classic greedy search algorithm can be viewed as solving the following optimization problem:
{\small
\begin{equation}\label{eq:greedy_step}
    \begin{aligned}\min_{\vb^{(t)}} \quad & L(Q(\vw,\vb^{(t)})\\\text{s.t.} \quad &\frac{1}{N}\sum_i^N b_i^{(t)} \le B, \\ 
    % &\sum_i^N |b_i^{(t)} -b_i^{(t-1)}| \le 1,\\
    & \|\vb^{(t)} - \vb^{(t-1)}\|_1 \le 1,\\
    % & |b_i^{(t)} -b_i^{(t-1)}| \le 1,\\ 
    & \|\vb^{(t)} - \vb^{(t-1)}\|_\infty \le 1, \quad\quad \vb \in \mathbb{Z}_{\ge 0} ^N. 
    \end{aligned}
\end{equation}
}
\vspace{-.15in}

The inefficiency comes from evaluating the objective function and the single-update enforced by the second constraint.

To obtain a scalable alternative, we introduce two relaxations. First, we replace the exact loss in the objective function with a first-order approximation based on sensitivity estimates. This relies on the observation that the greedy search procedure depends primarily on the \emph{relative ordering} of marginal gains, rather than their exact values. As shown in Section~3, our sensitivity estimate derived from quantized gradients reliably preserves this ordering in practice. Second, we relax the feasible set to allow batched updates, in which multiple blocks may change precision simultaneously. These relaxations yield the following approximate subproblem at iteration $t$:

\vspace{-.1in}
{\small
\begin{equation}\label{eq:relaxed_step}
    \begin{aligned}
    \min_{\vb^{(t)}} \quad & L^{(t-1)} - \sum_{i=1}^N (\vs_{\text{up}})_i\mathbb{I}(b_i^{(t)}= b_i^{(t-1)}+1) \\
    & \phantom{L^{(t-1)}} + \sum_{i=1}^N (\vs_{\text{down}})_i\mathbb{I}(b_i^{(t)} = b_i^{(t-1)}-1)\\
    \text{s.t.}\quad & \frac{1}{N} \sum_i^N b_i^{(t)} \le B, \\
    % &\sum_i^N |b_i^{(t)} -b_i^{(t-1)}| \le rN, \\
    & \|\vb^{(t)} - \vb^{(t-1)}\|_1 \le rN, \\
    % &|b_i^{(t)}-b_i^{(t-1)}|\le 1, \\ 
    & \|\vb^{(t)}- \vb^{(t-1)}\|_\infty \le 1, \quad\quad \vb \in \mathbb{Z}_{\ge 0} ^N.
    \end{aligned}
\end{equation}
}
\vspace{-.1in}

Here, $(\vs_{\text{up}})_i$ and $(\vs_{\text{down}})_i$ approximate the marginal loss change of increasing or decreasing the precision of block i, respectively, and $\gamma \in (0, 1]$ controls the batch size. Algorithm~2 implements this relaxed subproblem in iterations, incorporating several principled modifications to the classic greedy search to ensure scalability and numerical stability.

\begin{algorithm}[t]
\caption{Scalable Greedy Search}
\label{alg:sgreedy}
\begin{algorithmic}[1]
\STATE \textbf{Input:} weight partition $\Pi_{\vw}$, bit budget $B$, calibration dataset $\gD$, quantizer $Q(\cdot)$, update ratios $\gamma_0$ and $\gamma_{T}$. 
%\STATE \textbf{Output:} Bit allocation vector $\vb \in \mathbb{Z}_{\ge 0}^N$
\STATE initialize $b_i^{(0)} = \floor{B} \ \forall i$, $t = 0$, $k = \floor{\gamma_0 N}$
\WHILE{$k \ge \floor{\gamma_T N}$}
    \STATE sample $\gD^{(t)} \sim \gD$
    \STATE update $\vs_{\text{up}}^{(t)}$ and $\vs_{\text{down}}^{(t)}$ on $\gD^{(t)}$
    \IF{$\frac{1}{N}\sum_{i=1}^N b_i^{(t)} < B$}
    \STATE $\vb^{(t+1)} \leftarrow \vb^{(t)} + \sum_{i: s_i \in \mathrm{top}\text{-}k(\vs_{\text{up}}^{(t)})}\ve_{i}$
    \ELSE
    \STATE $\vb^{(t+1)} \leftarrow \vb^{(t)} + \sum_{i: s_i \in \mathrm{top}\text{-}\frac{k}{2}(\vs_{\text{up}}^{(t)})}\ve_{i}$
    \STATE $\vb^{(t+1)} \leftarrow \vb^{(t)} - \sum_{i: s_i \in \mathrm{bottom}\text{-}\frac{k}{2}(\vs_{\text{down}}^{(t)})}\ve_{i}$     
        \IF{{\small $L_{\gD^{(t)}}(Q(\vw, \vb^{(t+1)})) > L_{\gD^{(t)}}(Q(\vw, \vb^{(t)}))$}}
        \STATE $\vb^{(t+1)} \leftarrow \vb^{(t)}, k \leftarrow k/2$         
        \ENDIF
    \ENDIF
    \STATE $t \leftarrow t + 1$
\ENDWHILE
\STATE \textbf{Return} $\vb^{(t)}$
\end{algorithmic}
\end{algorithm}
\textbf{Two-stage batched updates.}
The optimal solution of the problem~(\ref{eq:relaxed_step}) admits two update rules, depending on whether the bit budget is active. If the current allocation under-utilizes the budget, i.e., $\frac{1}{N}\sum_i b_i^{(t)} < B$, we apply a pure expansion step by increasing the precision of the $k$ most sensitive blocks. When the budget is active, we perform a balanced update that increases the precision of $k/2$ most sensitive blocks by $(\vs_{\text{up}})_i$ and decreases the precision of the $k/2$ least sensitive blocks ranked by $(\vs_{\text{down}})_i$. Throughout both stages, batched updates substantially accelerate convergence while mitigating noise in the estimated gains.

\textbf{Acceptance checking.}
Since sensitivity estimates use sampled data and first-order approximations, noisy updates may increase loss. To ensure numerical stability, we include a lightweight acceptance check (line~11): if an update increases the loss on the current batch, it is rejected, and the batch size is reduced. This mechanism also serves as an implicit stopping criterion: as $k$ decreases, the algorithm converges to a solution that is locally optimal with respect to small exchange moves. 

\textbf{Warm-start.}
The classic greedy search starts from the minimal configuration; however, in our setting, $\vb^{(0)} = \vzero$ corresponds to a fully pruned model. More generally, when most components are heavily quantized into the extreme low-precision regime (e.g., 1-bit or pruned), the collapsed activations and unstable gradients lead to unreliable sensitivity estimates. To avoid this regime, we adopt a warm-start strategy by initializing the precision closer to the budget $\vb = \floor{B}$. While this initialization leaves some regions of the feasible space unexplored, the subsequent post-refinement stage enables effective local reallocation.

\begin{table*}[t]
    \centering
    \resizebox{0.98\textwidth}{!}{
    \begin{tabular}{c|c|c|c|cc|c|cc|c|cc}
        \toprule
        \multirow{2}{*}{Method} & \multirow{2}{*}{MP} & \multirow{2}{*}{bits} & \multicolumn{3}{c|}{Llama2 - 7B} & \multicolumn{3}{c|}{Llama3 - 8B} & \multicolumn{3}{c}{Llama3 - 70B} \\
         & & & Wiki2 $\downarrow$ & 0-Shot $\uparrow$ & MMLU $\uparrow$  & Wiki2 $\downarrow$ & 0-shot $\uparrow$ & MMLU $\uparrow$ & Wiki2 $\downarrow$ & 0-shot $\uparrow$ & MMLU $\uparrow$\\
         \midrule
         - & $\times$ & 16 & 5.12 & 66.55 & 45.86 & 5.54 & 70.69 & 65.34 & 2.59 & 76.88 & 78.74\\ 
         \midrule
         \myrowcolour%
         RTN-g128& $\times$ & 3.1 & 6.20 & 63.65 & 37.66 & 10.91 & 60.39 & 46.45 & 19.35 & 66.33 & 54.56\\
         GPTQ-g128 & $\times$ & 3.1 & 5.67 & 64.15 & 41.43 & 6.92 & 64.97 & 54.27 &  4.75 & 73.72 & 74.51 \\
         SlimLLM-g128 & $\checkmark$ & 3.1 & 5.81 & 63.06 & 39.79 & 6.85 & 66.03 & 55.03 & 4.08 & | & | \\ 
         GuidedQuant ($\star$) & $\times$ & 3.0 & 5.57 & 65.12 & 43.38 & 6.99 & \textbf{68.58} & 59.07 & 3.90  & \textbf{75.51} & 76.86 \\
         \myrowcolour% 
         \ours + RTN & $\checkmark$ & 3.1 & \textbf{5.52} & \textbf{65.53} & \textbf{43.48} & \textbf{6.63} & 67.87 & \textbf{59.35} & \textbf{3.69} & 75.00 & \textbf{76.88} \\
         \midrule
         \myrowcolour% 
         RTN-g128& $\times$ & 2.1 & 5e3 & 35.80& 24.80& 3e5 & 37.16 & 24.10 & 2e4 & 35.61 & 23.02\\
         GPTQ-g128 & $\times$ & 2.1 & 16.69 & 40.67 & 24.95 & 340 & 35.82 & 25.47 & 28.80 & 46.60 & 26.63 \\
         SlimLLM-g128 & $\checkmark$ & 2.1 & 15.38 & 47.00 & 23.69 & 66.06 & 37.13 & 25.51 & 9.46 & | & | \\
         GuidedQuant ($\star$) & $\times$ & 2.0 & 8.83 & 51.35 & 29.37 & 30.80 & 56.21 & 26.97 & 10.21 & 64.39 & 45.55\\
         \myrowcolour% 
         \ours + RTN & $\checkmark$ & 2.1 & \textbf{7.49} & \textbf{59.53} & \textbf{34.01} & \textbf{11.80} & \textbf{58.53} & \textbf{39.36} & \textbf{7.80} & \textbf{64.72} & \textbf{58.82} \\
         \bottomrule
    \end{tabular}
    }
    \vspace{-.05in}
    \caption{Evaluation of quantized LLMs under 2-3 bit regime. Wiki2 reports perplexity on WikiText-2 (context length = 4096 for Llama2, 8192 for Llama3). 0-shot denotes average accuracy over six zero-shot tasks, and MMLU reports 5-shot accuracy. Methods marked with $\star$ use non-uniform quantizers; all others use uniform quantizers. MP indicates mixed-precision.}
    \vspace{-.1in}
    \label{tab:main_result}
\end{table*}
\section{Experiments}
\textbf{Implementation.}
We integrate \ours with rounding-to-nearest (RTN) uniform scalar quantizer (group size 128). After channel reordering, each weight matrix is partitioned into blocks with size $64 \times 128$. The greedy search runs with $\gamma_0 = 5\%$ and $\gamma_T = 2\%$, and search space $\mathcal{B}=\{1,2,\ldots,8\}$. Sensitivity estimation uses 128 sequences of 4096 tokens from RedPajama \cite{weber2024redpajama}.

\textbf{Models and Benchmark.}
We evaluate our method on a diverse set of open-source LLMs from the Llama-2/3 \cite{touvron2023llama2,grattafiori2024llama3}, and Gemma-2 \cite{team2024gemma} families, covering base and instruction-tuned variants up to 70B parameters. Following prior quantization work, we report perplexity scores on the WikiText-2 \citep{merity2016wiki} test set and zero-shot accuracies on WinoGrande \citep{sakaguchi2021winogrande}, PiQA \citep{bisk2020piqa}, HellaSwag \cite{zellers2019hellaswag}, ARC-easy, ARC-challenge \citep{clark2018arc}, and BoolQ \citep{clark2019boolq}, 
together with 5-shot accuracy on MMLU \citep{hendryckstest2021mmlu}%. , to assess general language understanding and representation robustness. 
. For instruction-tuned models, we evaluate performance degradation on more challenging math and coding benchmarks, including GSM8K \citep{cobbe2021gsm8k} and MBPP \citep{austin2021mbpp}. All task accuracies are evaluated using the LM Evaluation Harness \citep{eval-harness}.

\textbf{Baseline Methods.}
We compare against four representative post-training weight-only
quantization methods. RTN serves as a naive uniform-precision baseline. GPTQ \cite{frantar2023gptq} and GuidedQuant \cite{kim2025guidedquant} represent
state-of-the-art sensitivity-aware uniform-precision methods, using uniform and non-uniform quantizers, respectively. SlimLLM \cite{huang2024slim}
is a mixed-precision quantization method. Our evaluation focuses on scalar quantization schemes. Although recent vector-quantization-based approaches \cite{egiazarian2024aqlm, tseng2024quip, tseng2024qtip} achieve higher accuracies, they typically incur additional runtime overhead due to large codebook lookups, on-the-fly codebook generation, or extra matrix multiplications during dequantization. Importantly, \ours is orthogonal to the choice of quantizer and can be readily combined with advanced quantization schemes, including vector quantization.

\subsection{Main Results}
Across all evaluated models and bit budgets, ScaleBITS consistently improves the accuracy–compression tradeoff over both uniform-precision and prior mixed-precision baselines. 
Table~\ref{tab:main_result} presents results on LLaMA-2 and 3 base models, with extended results on instruction-tuned variants and other architectures in Appendix~\ref{app:results}.

\textbf{Comparison with uniform-precision quantization.}
 To isolate the benefit of mixed-precision, we first compare \ours with its uniform-precision RTN backend. Our mixed-precision framework significantly improves RTN across all tasks.
 For example, at an average budget of ~2 bits on LLaMA-3-70B, mixed-precision RTN improves zero-shot accuracy from 35.61\% to 64.72\% and MMLU accuracy from 23.02\% to 58.82\%.

We next compare \ours against strong sensitivity-aware methods, including GPTQ (with activation reordering) and GuidedQuant.
Despite their more sophisticated error modeling and, in GuidedQuant's case, a non-uniform quantizer (LNQ), \ours with a simple RTN backend consistently achieves better or comparable accuracies, especially at the 2-bit regime.  This indicates that proper bit allocation provides larger gains than increasingly refined quantization schemes within a fixed precision. 

\textbf{Comparison with mixed-precision quantization.}
Finally, we compare \ours with SlimLLM, the state-of-the-art mixed-precision approach built on GPTQ; additional baselines are reported in Appendix~\ref{app:results}. In addition to achieving better model quality at matched budgets, a key advantage of \ours is its ability to realize a wide range of target model sizes through global precision allocation. Uniform-precision quantization provides only a few discrete operating points. %, since all weights share the same bitwidth. 
SlimLLM relaxes this constraint by allowing limited variation within each layer, but restricts bitwidth choices to three neighboring values and enforces a balanced ratio inside each layer. As a result, its achievable model sizes are still discrete, and its precision decisions are largely confined within individual layers.

In contrast, \ours performs flexible precision allocation across all weight blocks, enabling direct control over the global bit budget. Figure~\ref{fig:accuracy} illustrates the resulting bitwidth–perplexity curve achieved by \ours, revealing a dense set of achievable tradeoff points that are inaccessible to both uniform-precision methods and SlimLLM.

\subsection{Ablation and Analysis}\label{sec:ablation}
\begin{figure}[h]
    \centering
    \includegraphics[width=0.45\linewidth]{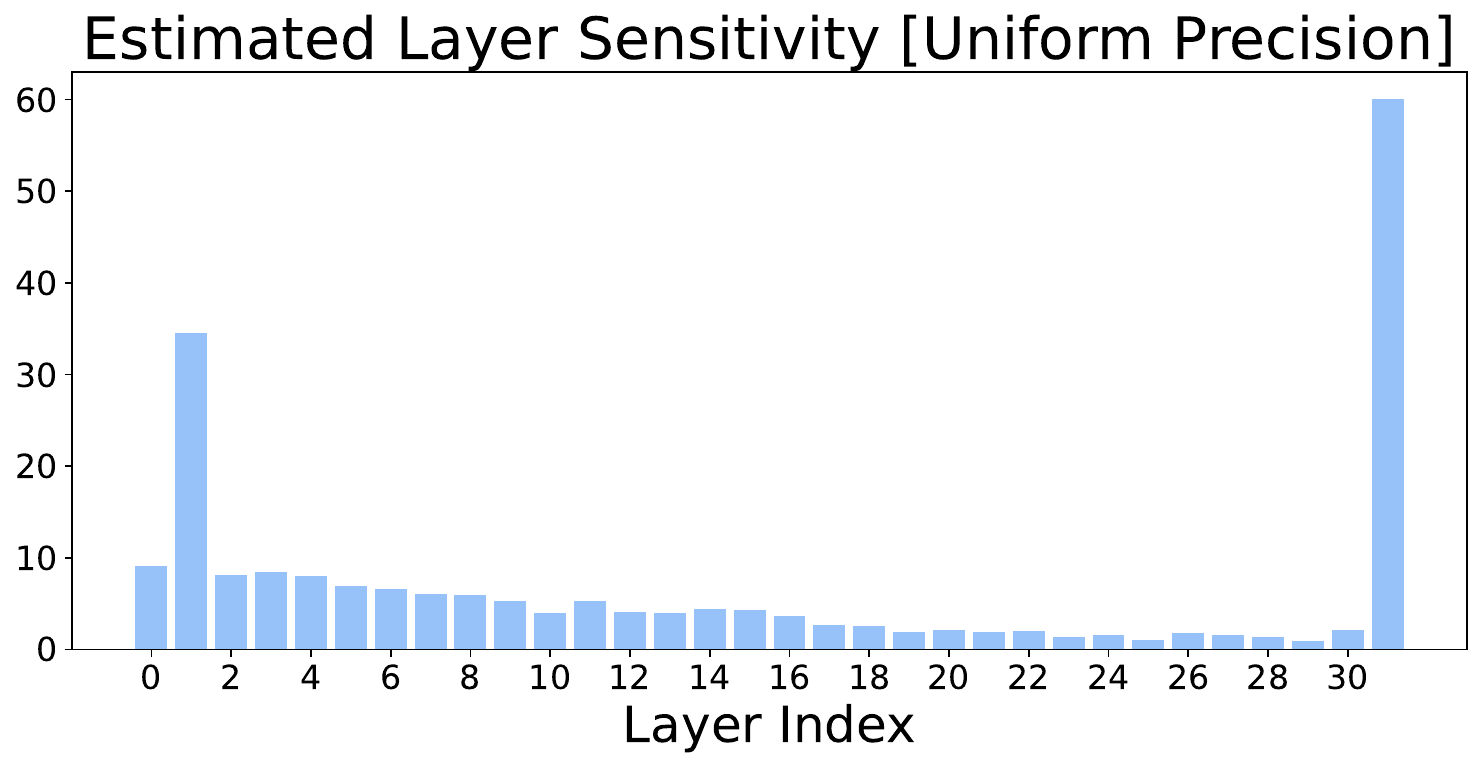}
    \hspace{0.02\linewidth}
    \includegraphics[width=0.45\linewidth]{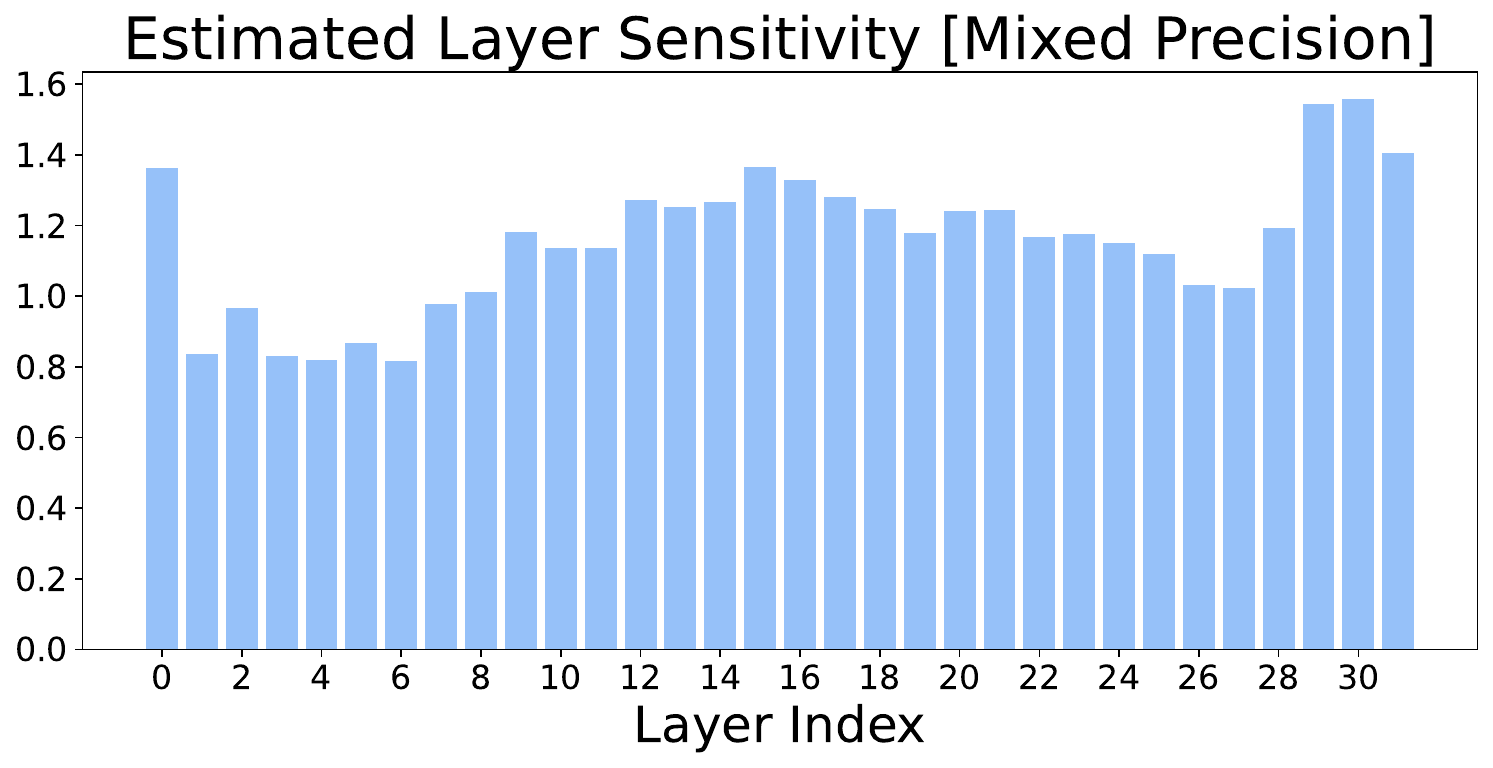}
    \vspace{-.1in}
    \caption{The estimated layer sensitivity under (left) uniform-precision and (right) mixed-precision quantization.}
    \label{fig:itr_sensitivity}
\end{figure}

\begin{figure}[h]
    \centering
    \includegraphics[width=0.45\linewidth]{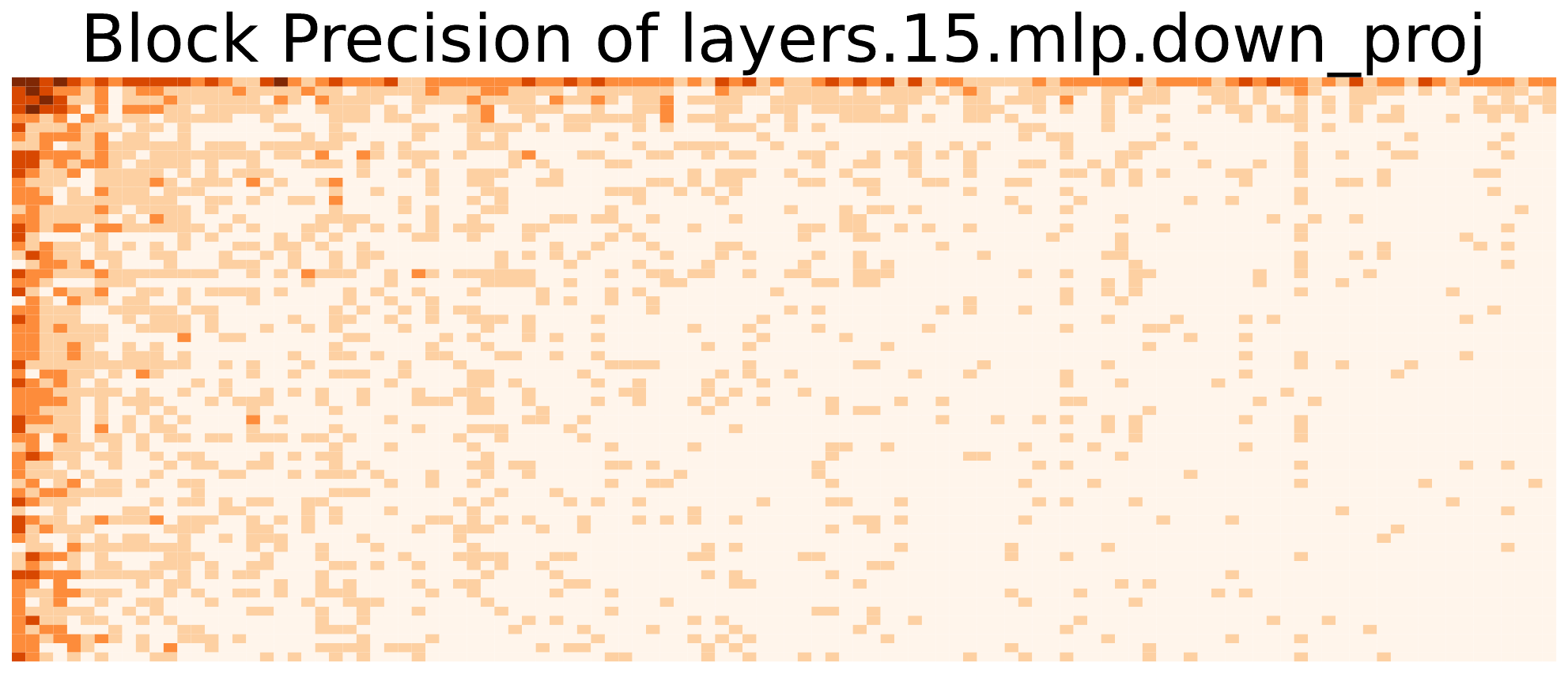} 
    \hspace{0.03\linewidth}
    \includegraphics[width=0.45\linewidth]{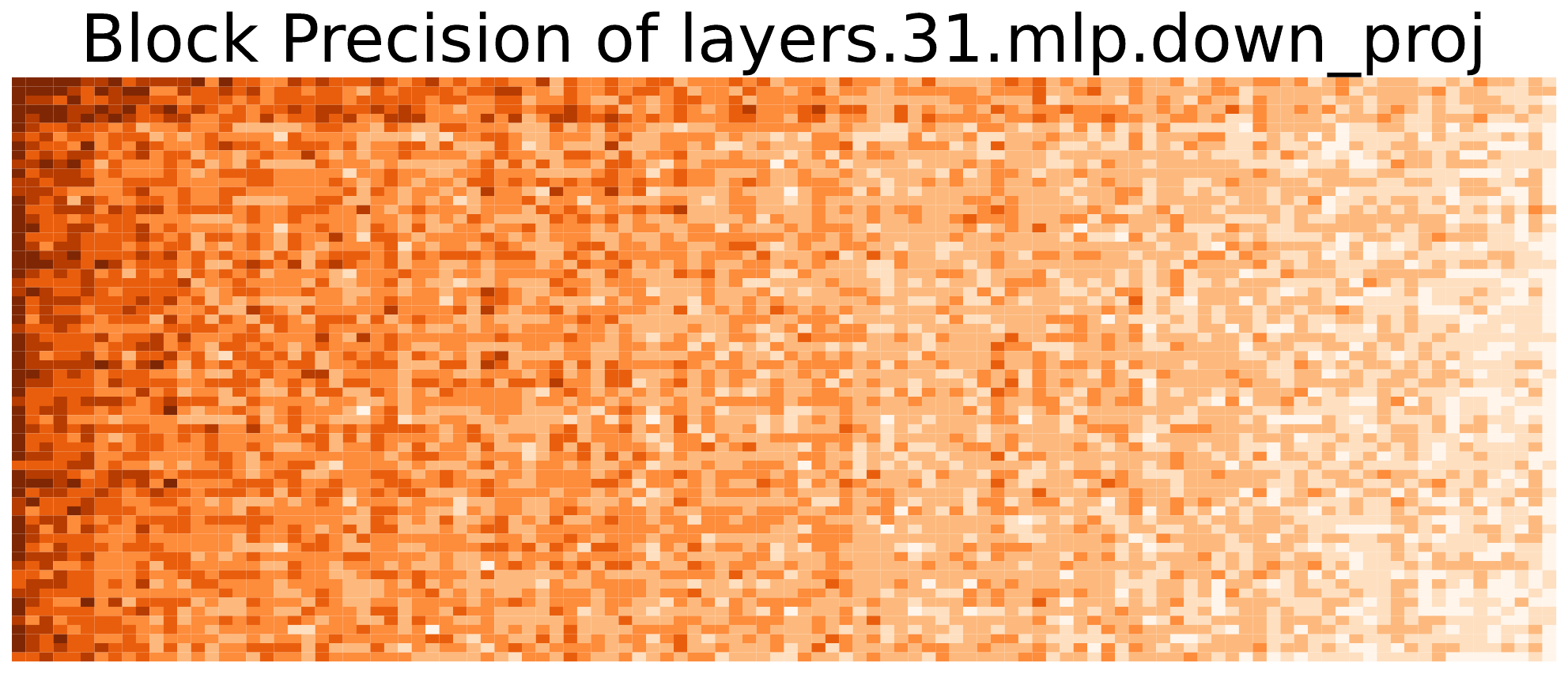}
    \vspace{-.1in}
    \caption{Block precision allocation of (left) a middle and (right) the last \texttt{down\_proj} layers in Llama3-8B model.}
    \label{fig:block_precision}
    \vspace{-.05in}
\end{figure}

We analyze the behavior of \ours using 3-bit Llama3-8B quantization as a case study. Figure~\ref{fig:itr_sensitivity} compares the estimated layer-wise sensitivity before (single precision) and after (learned mixed-precision) applying our precision search, where sensitivity reflects each layer’s contribution to end-to-end loss degradation. Starting from the uniform-precision initialization, sensitivity is highly concentrated in the second and last layers; after mixed-precision allocation, \ours substantially reduces these peaks, resulting in a more balanced sensitivity distribution across the model. Figure~\ref{fig:block_precision} visualizes the learned block-wise precision allocation for two representative layers. Within each matrix, blocks near the top-left corner—corresponding to more sensitive weights—are assigned higher precision (darker color), confirming that the learned allocation aligns with our intended design. Moreover, the two layers receive noticeably different total bit budgets, highlighting that our precision allocation is performed globally across layers. Additional ablation studies on channel reordering, sensitivity estimates, and hyperparameters are provided in Appendix~\ref{sec:ablation}.


\subsection{Efficiency}
\textbf{Precision Search.}
To demonstrate the efficiency of Algorithm~\ref{alg:sgreedy}, we report the end-to-end quantization time (including precision search) and iterations required to quantize Llama3-8B model to 3-4 bit regime. Table~\ref{tab:time} shows \ours achieves a $3-6\times$ speedup over SlimLLM, despite performing a global search over an exponentially larger precision space. The search time of \ours varies with the distance to the target bit budget B, whereas SlimLLM does not support arbitrary budget optimization.

\begin{table}[h]
    \centering
    \resizebox{0.4\textwidth}{!}{
    \begin{tabular}{c|c|c}
        \toprule
         Method & Time (h) & Search Iterations \\
         \midrule
         SlimLLM & $3$ & |\\
         Classic Greedy Search & $\approx 10^{10}$ &  $\approx 3 \times 10^6$\\
         \myrowcolour
         \ours & $0.5 - 1$ & $16 - 36$ \\
         \bottomrule
    \end{tabular}
    }
    \vspace{-.05in}
    \caption{Time required to quantize LLaMA-3-8B with mixed-precision schemes on a single NVIDIA H100 GPU.}
    \label{tab:time}
\end{table}
\textbf{Inference Kernel.}
To demonstrate the hardware efficiency and practical feasibility of our block-wise weight partition, we implement a lightweight Triton kernel, fusing mixed-precision dequantization with matrix multiplication. Dequantization is performed at the same block granularity as the GEMM tiling, with uniform bitwidth per tile. 
As a result, mixed-precision tiles can be processed uniformly and directly fed into Tensor Core–accelerated matrix multiplication without introducing warp-level divergence or requiring separate kernels for different precisions.

Table~\ref{tab:kernel} reports the end-to-end latency of small-batch matrix multiplication under different precision mixtures. The results show that mixed-precision quantization incurs no measurable latency overhead compared to uniform-precision quantization at the same average bitwidth.
For reference, the BF16 baseline uses a fully optimized CUTLASS kernel. Our Triton kernel is intended as a proof-of-concept rather than a fully optimized implementation; nevertheless, the results already indicate that block-wise mixed-precision can be realized efficiently on modern accelerators.
\begin{table}[h]
    \centering
    \resizebox{0.35\textwidth}{!}{
    \begin{tabular}{c|c|c|c}
    \toprule
        \multirow{2}{*}{Bits} & MP Ratio & \multicolumn{2}{c}{Latency (us)} \\
        & [INT2, INT4, INT8] & BS=16 & BS=32 \\
    \midrule
         4 & [0, 100\%, 0] & 57.8 & 65.0\\
         4 & [40\%, 40\%, 20\%] & 58.0 & 65.9 \\
    \midrule
        16 & | & \multicolumn{2}{c}{86.0} \\
    \bottomrule
    \end{tabular}
    }
    \vspace{-.05in}
    \caption{Kernel latency under mixed-precision (MP) settings for an $8192 \times 8192$ LLM-scale projection (BS=batch size).}
    \label{tab:kernel}
    \vspace{-.1in}
\end{table}

\section{Conclusion}
In this paper, we introduce \ours, a hardware-aligned mixed-precision weight quantization framework for LLMs, powered by structured weight partitioning and automated precision search. By exploiting the dynamic nature and spatial structure of weight sensitivity, \ours enables fine-grained control of the global bit budget and achieves superior perplexity–compression tradeoffs in the ultra-low-bit regime while preserving hardware efficiency. Future directions include extending the framework to joint weight–activation precision allocation and incorporating lightweight fine-tuning or training-time adaptation to further improve robustness under extreme compression.


\newpage
\section*{Impact Statement}
This work introduces a hardware-aligned mixed-precision quantization framework that improves the efficiency of large language model (LLM) deployment. By reducing memory usage and data movement, our method can decrease the environmental footprint of large-scale AI systems and make LLMs more accessible in resource-constrained settings. At the same time, improved efficiency may enable wider deployment of LLMs, potentially amplifying concerns about misuse, misinformation, and bias propagation. Although our quantization method does not directly modify model behavior, responsible deployment remains essential. We call for continued research on how compression techniques interact with model robustness, fairness, and safety.

\bibliography{references}
\bibliographystyle{icml2025}

%%%%%%%%%%%%%%%%%%%%%%%%%%%%%%%%%%%%%%%%%%%%%%%%%%%%%%%%%%%%%%%%%%%%%%%%%%%%%%%
%%%%%%%%%%%%%%%%%%%%%%%%%%%%%%%%%%%%%%%%%%%%%%%%%%%%%%%%%%%%%%%%%%%%%%%%%%%%%%%
% APPENDIX
%%%%%%%%%%%%%%%%%%%%%%%%%%%%%%%%%%%%%%%%%%%%%%%%%%%%%%%%%%%%%%%%%%%%%%%%%%%%%%%
%%%%%%%%%%%%%%%%%%%%%%%%%%%%%%%%%%%%%%%%%%%%%%%%%%%%%%%%%%%%%%%%%%%%%%%%%%%%%%%
\newpage
\appendix
\onecolumn
\section*{Appendix.}
% You can have as much text here as you want. The main body must be at most $8$ pages long.
% For the final version, one more page can be added.
% If you want, you can use an appendix like this one.  

% The $\mathtt{\backslash onecolumn}$ command above can be kept in place if you prefer a one-column appendix, or can be removed if you prefer a two-column appendix.  Apart from this possible change, the style (font size, spacing, margins, page numbering, etc.) should be kept the same as the main body.
In this appendix, we provide further details as follows:
\begin{itemize}
\renewcommand{\labelitemi}{}
    \item \ref{app:related}. \hyperref[app:related]{Related Work}
    \item \ref{app:assumption}. \hyperref[app:assumption]{Assumptions in Problem Formulation}
    \item \ref{app:sensitivity}. \hyperref[app:sensitivity]{Weight Sensitivity}
    \item \ref{app:reorder}. \hyperref[app:reorder]{Channel Reordering}
    \item \ref{app:ablation}. \hyperref[app:ablation]{Ablation Study}
    \item \ref{app:visual}. \hyperref[app:visual]{Other Visualization and Analysis}
    \item \ref{app:results}. \hyperref[app:results]{Other Experiment Results}
\end{itemize}

\section{Related Work}\label{app:related}
\subsection{Post Training Weight Quantization}
Model quantization has been widely studied, with early work focusing primarily on convolutional neural networks (CNNs). The emergence of LLMs has driven renewed interest in efficient post-training quantization (PTQ) methods that reduce memory footprint and inference cost without retraining. A large body of work focuses on weight-only quantization, where all weights share the same bitwidth by default. Early LLM PTQ methods focused on improving uniform scalar quantizers, which are valued for their simple implementation and efficient inference-time decoding. Representative approaches include AWQ \cite{lin2024awq}, GPTQ \cite{frantar2023gptq}, and OmniQuant \cite{shao2024omniquant}. While these methods perform well at moderate precision (e.g., 4 bits), they exhibit significant accuracy degradation in the ultra-low-bit regime (below 4 bits). To better capture the wide dynamic range of weights with limited quantization levels, SqueezeLLM \cite{kim2023squeezellm}, and GuidedQuant \cite{kim2025guidedquant} explored non-uniform scalar quantization. 

More recent methods, including AQLM \cite{egiazarian2024aqlm}, Quip\# \cite{tseng2024quip}, and QTIP \cite{tseng2024qtip}, investigate vector quantization, while ICQuant \cite{li2025icquant} proposes separately quantizing outlier weights and compressing their storage through index-coding. These advances substantially improve model quality at very low bitrates. However, compared to standard scalar quantization, they introduce more complex decoding procedures and memory access patterns, making them less suitable for hardware-efficient, latency-sensitive deployment.

\subsection{Weight Sensitivity}
It has been widely observed across a variety of LLM architectures that weight importance is highly non-uniform \cite{dettmers2023spqr, kim2023squeezellm, sunmassive}. In particular, \cite{lin2024awq} shows that keeping 0.1–1\% of important weights in full precision can significantly improve quantization quality; while \cite{super-weight} demonstrates that removing a single weight element can severely disrupt model behavior. Sensitivity estimation is therefore a central component of model quantization and pruning methods. A range of sensitivity metrics have been explored, including magnitude-based criteria \cite{yuan2024pbllm}, first-order gradient information \cite{li2023llm}, and second-order approximations \cite{frantar2023gptq,kim2023squeezellm, kim2025guidedquant}. However, as discussed in Section~\ref{sec:sensitivity}, most existing approaches estimate sensitivity at the full-precision model and treat it as a static property to guide quantization. Our work revisits this assumption and shows that weight sensitivity is inherently dynamic and that adaptively measuring it can better capture the model’s internal dependency.

\subsection{Mixed-Precision Quantization}
Mixed-precision quantization addresses the non-uniform sensitivity by assigning different bitwidths to different weight components. Existing methods mainly differ along two axes: (i) partition granularity and (ii) precision allocation strategy.

\textbf{Partition Granularity.}
Early mixed-precision methods such as HAQ \cite{wang2019haq} and HAWQ \cite{dong2020hawq} adopt coarse layer-wise precision assignment, which limits flexibility when sensitivity varies significantly within layers.
To improve flexibility, many recent approaches for LLMs explore fine-grained schemes. SpQR \cite{dettmers2023spqr}, SqueezeLLM \cite{kim2023squeezellm}, detect a small fraction ($\sim 0.5\%$) of the most sensitive weights and keep them in full precision while aggressively quantizing the rest to 2-4 bits. PB-LLM \cite{yuan2024pbllm} keeps up to 30\% sensitive weights in higher precision while binarizing the remaining. These methods improve accuracy by focusing precision on a small subset of weights, but the resulting irregular sparsity introduces index-storage overhead and leads to inefficient memory access and execution patterns on modern accelerators \cite{li2024fast}.

Between these two extremes, structured intermediate granularities have been proposed. OWQ \cite{lee2024owq} identifies weak weight columns and stores them in full precision, and SlimLLM \cite{huang2024slim} groups contiguous columns and assigns different bits accordingly. These structured partitions reduce overhead, but their grouping boundaries are determined by the original matrix structure and therefore do not fully capture the distribution of weight sensitivity.

\textbf{Precision Allocation Strategy.}
Early work often formulates this as a search problem solved via reinforcement learning, evolutionary algorithms, or differentiable optimization \cite{wang2019haq,cai2020rethinking,yuan2020evoq}. For example, the classic greedy algorithm (Algorithm~\ref{alg:sgreedy}) has been applied in coarse layer-wise precision search for small CNNs \cite{chen2021towards}. While effective at a small scale, these approaches become prohibitively expensive for LLMs.

Consequently, many LLM-oriented methods rely on heuristics or highly constrained allocation strategies \cite{rakka2025mixed}. Most existing schemes \cite{kim2023squeezellm,dettmers2023spqr,yuan2024pbllm,guan2024aptq,lee2024owq} restrict the search space to two available precision choices and select a fixed fraction of weights to remain in higher precision, based on sensitivity scores estimated once at the full-precision model. SlimLLM \cite{huang2024slim} performs a restricted layer-wise search over a small set of neighboring bitwidths (b-1, b, b+1) to match a target average precision, but precision allocation across layers remains unexplored. These design choices reduce search complexity but also restrict the achievable accuracy-compression trade-offs. Recent efforts such as AMQ \cite{lee2025amq} and \cite{boudouh2025constrained} begin to explore automated precision search for LLMs, but remain focused on coarse, layer-level decisions and do not extend to fine-grained, such as block-wise partitions.

Finally, a related but orthogonal line of work studies adaptive-precision models that support multiple deployment settings (e.g., Any-Precision LLM \cite{park2024anyprecision}) or vary precision across decoding phases \cite{chen2025progressive}. These methods address dynamic deployment scenarios rather than static, budget-constrained mixed-precision allocation.

\vspace{.1in}
\begin{algorithm}[h]
\caption{Classic Greedy Search}
\label{alg:cgreedy}
\begin{algorithmic}[1]
\STATE \textbf{Input:} Weight partition $\Pi_{\vw}$, bit budget $B$, calibration dataset $\gD$, quantizer $Q(\cdot)$
%\STATE \textbf{Output:} Bit allocation vector $\vb \in \mathbb{Z}_{\ge 0}^N$
\STATE Initialize $\vb^{(0)} = \mathbf{0}$, $t = 0$
\WHILE{$\frac{1}{N}\sum_{i=1}^N b_i^{(t)} < B$}
    \FOR{$i = 1,\dots,N$}
        \STATE $\Delta_i = L_{\gD}(Q(\vw,\vb^{(t)})-L_{\gD}(Q(\vw, \vb^{(t)} + \ve_i))$
    \ENDFOR
    \STATE $\vb^{(t+1)} \leftarrow \vb^{(t)} + \ve_{i^\star},$ with $i^\star = \arg\max_i \Delta_i.$ 
    \STATE $t \leftarrow t + 1$
\ENDWHILE
\STATE \textbf{Return} $\vb^{(t)}$
\end{algorithmic}
\end{algorithm}

\newpage
\section{Assumptions in Problem Formulation}\label{app:assumption}
\begin{figure}[h]
    \centering
    \includegraphics[width=0.35\linewidth]{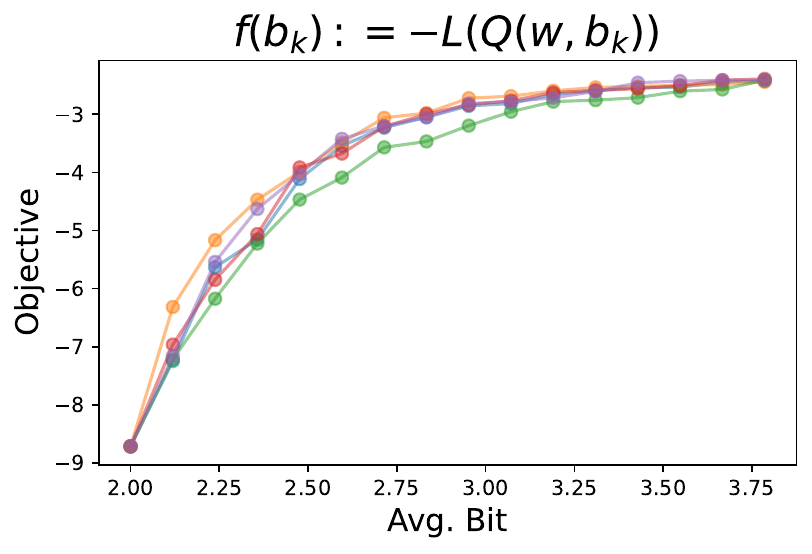}
    \hspace{0.05\linewidth}
    \includegraphics[width=0.35\linewidth]{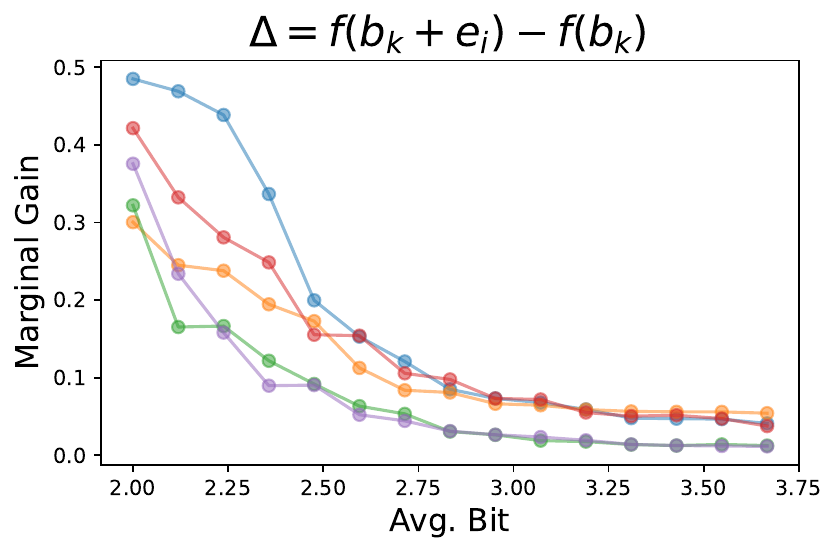}
    \caption{Empirical verification of monotonicity and diminishing returns under layer-wise precision allocation. Left: the objective value $f(\vb)$ improves monotonically with increasing average precision. Right: the marginal gain of adding one bit to a fixed component decreases as the model becomes more precise.}
    \label{fig:assumption}
\end{figure}

Given a weight partition $\Pi_\vw$ and a total memory budget $B$ (bits per weight), recall that we determine the precision allocation by solving the following optimization problem
\begin{equation}
    \begin{aligned}\min_\vb \quad & L(Q(\vw,\vb))\\\text{s.t.} \quad &\frac{1}{N}\sum_i^N b_i \le B, \\ & \vb \in \mathbb{Z}_{\ge 0} ^N, \end{aligned}
\end{equation}
where $Q(\vw,\vb)$ denotes the quantized model under precision assignment $\vb$. Let $f(b) := -L(Q(w, b))$. The problem is equivalent to maximizing $f(\vb)$ under a knapsack constraint on the integer lattice. To connect this formulation to classic greedy optimization results, we introduce two assumptions that are commonly used in the analysis of discrete resource allocation problems. We emphasize that our proposed algorithm does not rely on these assumptions for correctness or implementation; they are used solely to motivate and interpret the greedy-style optimization framework.
\begin{itemize}
    \item (Monotonicity) $f$ is monotone over the integer lattice: for any $\vb' \ge \vb$, $$f(\vb') \ge f(\vb).$$
    This reflects a natural property that increasing the precision of any component cannot worsen the loss.
    \item (DR-submodularity) $f$ is \textit{diminishing-returns submodular} \citep{soma2015generalization}: for all $\vb' \le \vb$ and coordinates $i$, $$f(\vb+\ve_i) - f(\vb) \le f(\vb' + \ve_i) - f(\vb'),$$ 
    where $\ve_i$ is the $i$-th standard basis vector. Intuitively, this implies that the marginal benefit of increasing precision for a component decreases as the overall model becomes more precise.
\end{itemize}
Under these assumptions, the mixed-precision allocation problem admits a classic greedy solution with approximation guarantees. We next empirically examine these assumptions through a simple sanity-check experiment, using a small calibration dataset with 128 sequences of 2048 tokens. For evaluation efficiency, we adopt a coarse layer-wise partition. In each trial, we randomly sample a coordinate update $\ve_i$ and randomly generate a monotone sequence of precision vectors $\{\vb^{k}\}_{k=1}^K$ with average precision ranging from 2 to 4 bits, such that $\vb_1 < \vb_2 < \ldots < \vb_N$. Figure~\ref{fig:assumption} plots the objective values $f(\vb_k)$ and the marginal gains of adding one bit to a fixed component $\Delta_k=f(\vb_k + \ve_i) - f(\vb_k)$ for five independently sampled trials. Both exhibit monotonically increasing (or decreasing) patterns, indicating that monotonicity and diminishing returns hold approximately in practice.

% Lemma~\ref{lm:mono} shows that the monotonicity is guaranteed when the quantizer is good enough.
% \begin{lemma}\label{lm:mono}
% Suppose $Q^*(\cdot, \cdot)$ is an optimal quantizer such that 
% \begin{equation}
%     Q^*(\vw, \vb) = \argmin_{\tilde{\vw}\in \gW(\vb)} L(\tilde{\vw}),
% \end{equation}
% where $\gW(\vb)$ denotes the set of weights representable under precision $\vb$. Then $f(\vb)$ is monotonically non-decreasing with respect to $\vb$.
% \end{lemma}
% \begin{assumption}[Monotonicity]
% $f$ is monotone over the integer lattice: for any $\vb' \ge \vb$,
% $$f(\vb') \ge f(\vb).$$
% \end{assumption}

% \begin{assumption}[DR-submodularity]
% $f$ is \textit{diminishing-returns submodular} \citep{soma2015generalization}: for all $\vb' \le \vb$ and every coordinates $i$, 
% $$
% f(\vb+\ve_i) - f(\vb) \le f(\vb' + \ve_i) - f(\vb'),
% $$
% where $\ve_i$ is the $i$-th standard basis vector.
% \end{assumption}

\section{Weight Sensitivity}\label{app:sensitivity}
\subsection{Sensitivity Estimation}
\begin{figure}[h]
    \vspace{.05in}
    \centering
    \includegraphics[width=0.45\linewidth]{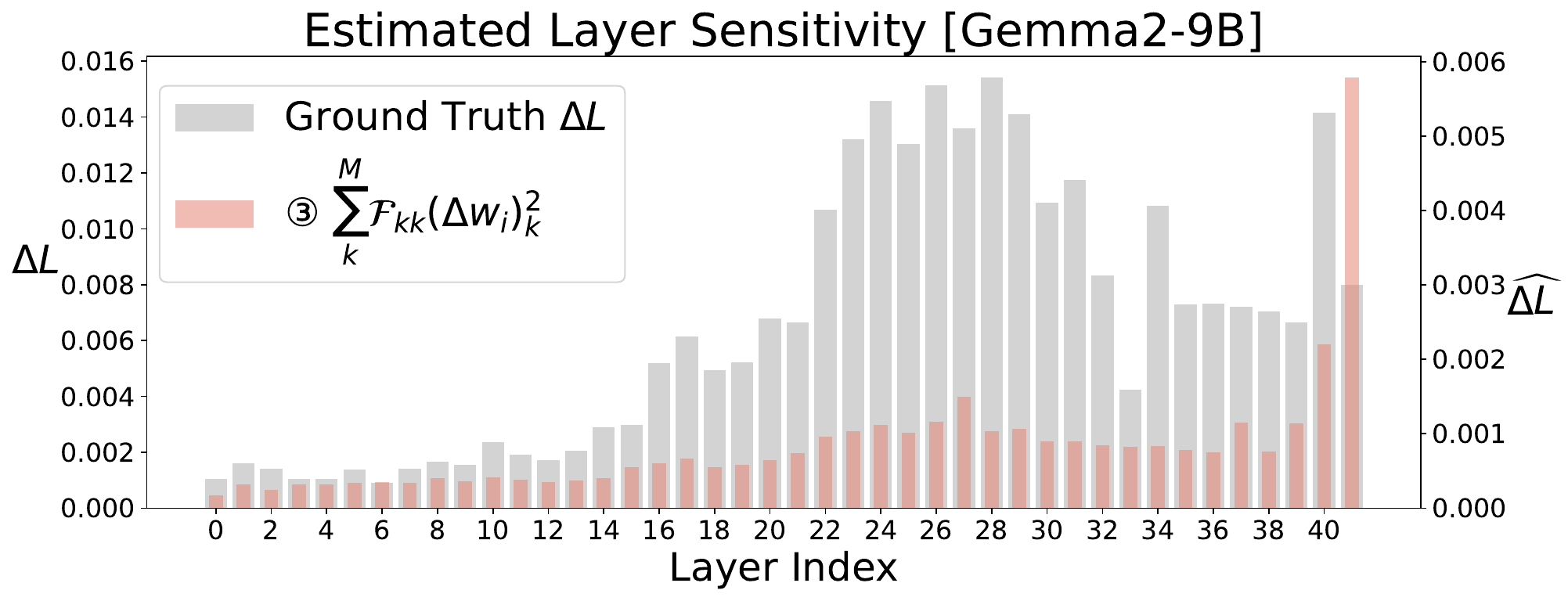}
    \includegraphics[width=0.45\linewidth]{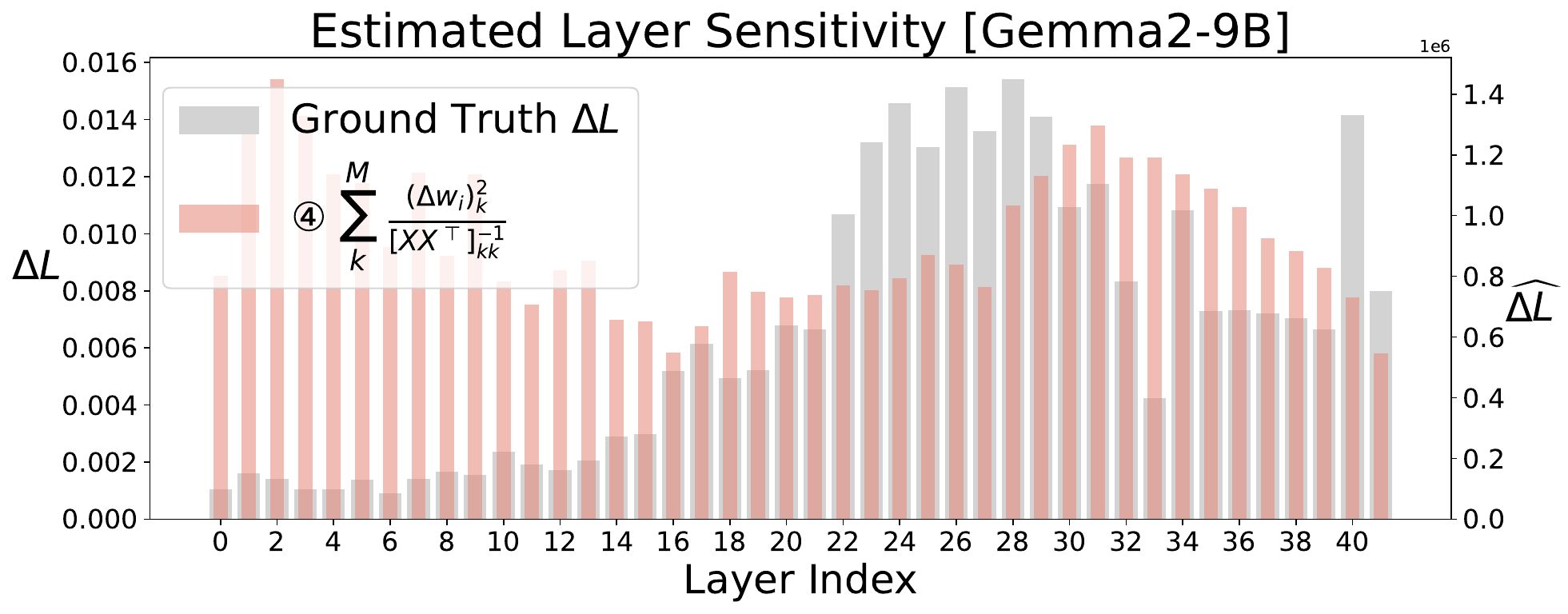}
    \vspace{-.1in}
    \caption{Estimated layer sensitivity of Gemma2-9B using existing second-order based metrics.}
    \label{fig:layer_sensitivity_more}
\end{figure}

In Figure~\ref{fig:layer_sensitivity} (Section~\ref{sec:sensitivity}), we compare estimated layer-wise sensitivities using two first-order metrics (ours in Eq.~\ref{eq:sensitivity} and \circled{1} in Table~\ref{tab:sensitivity}). Figure~\ref{fig:layer_sensitivity_more} reports the corresponding results for the two second-order metrics (\circled{3} and \circled{4} in Table~\ref{tab:sensitivity}), both of which fail to recover the correct sensitivity ranking.

The ground-truth layer sensitivity is computed as the loss change incurred by restoring a single layer from INT3-g128 to BF16 in an otherwise quantized model. We adopt this setting instead of quantizing one layer at a time, as it better reflects the mixed-precision objective, where most weights are quantized. This result also highlights that sensitivity is inherently dynamic: derivative information measured at the full-precision model no longer accurately characterize the loss landscape after aggressive quantization.

\begin{figure}[h]
    \vspace{.05in}
    \centering
    \includegraphics[width=0.45\linewidth]{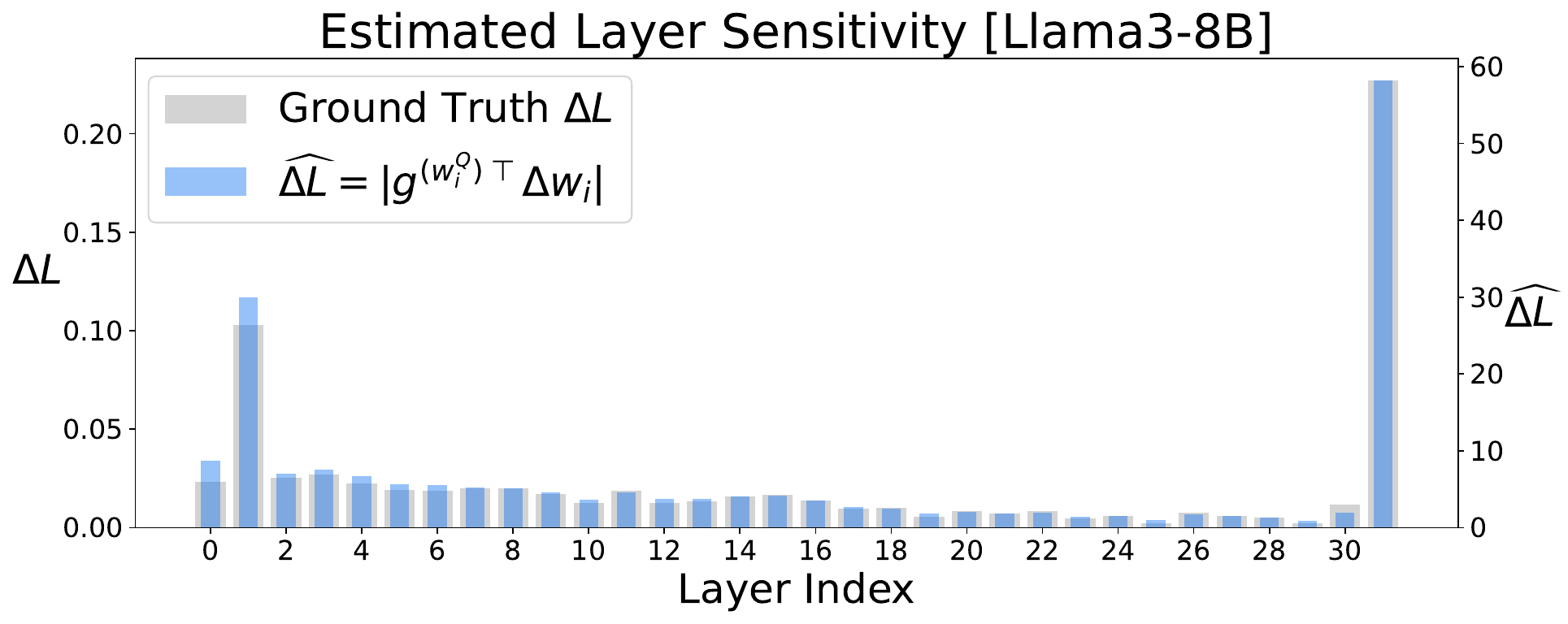}
    \includegraphics[width=0.45\linewidth]{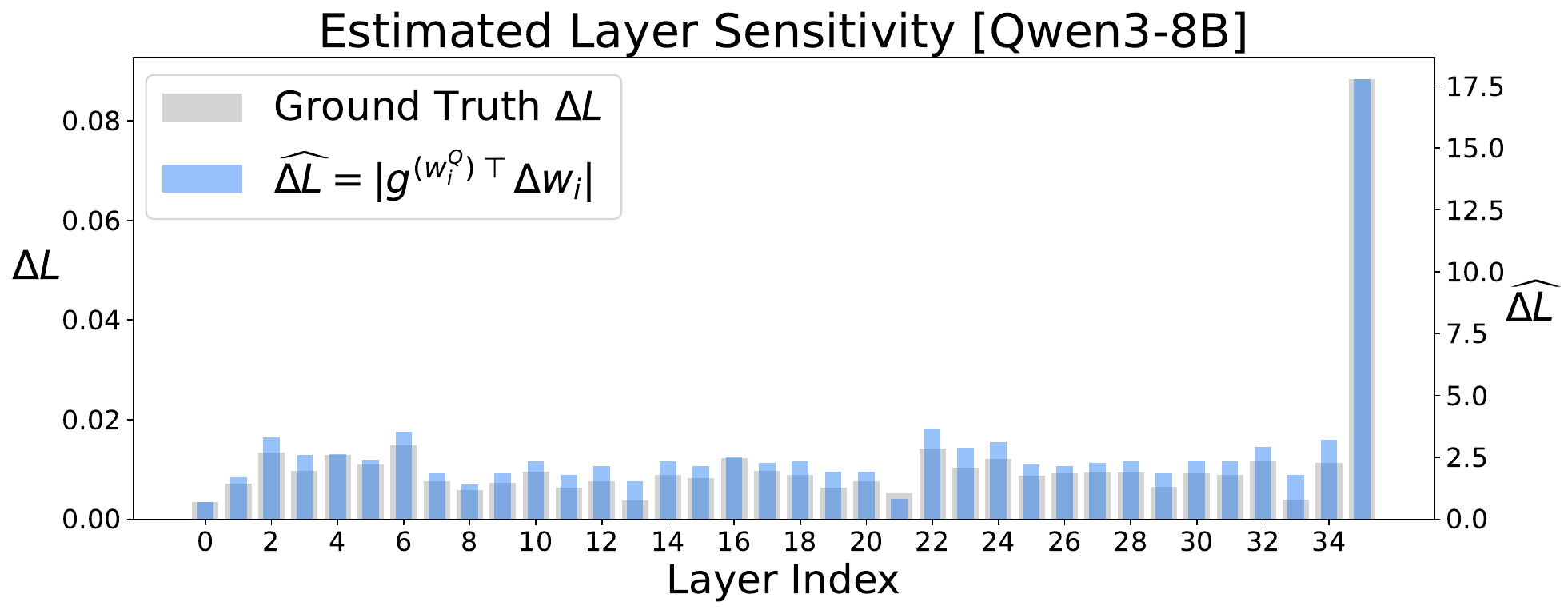}
    \vspace{-.1in}
    \caption{Estimated layer sensitivity of Llama3-8B and Qwen3-8B using our new sensitivity metric.}
    \label{fig:layer_sensitivity_lq}
\end{figure}

We further assess the generalizability of our sensitivity metric across different LLM families. As shown in Figure~\ref{fig:layer_sensitivity_lq}, Llama3 and Qwen3 models exhibit very different layer-wise sensitivity patterns compared with the Gemma2 model, with dominant (abnormally sensitive) layers appearing near the beginning and/or end. Despite these variations, our metric consistently captures the correct relative order of the layer sensitivities across all models.

\begin{figure}[h]
    \centering
\includegraphics[width=0.38\linewidth]{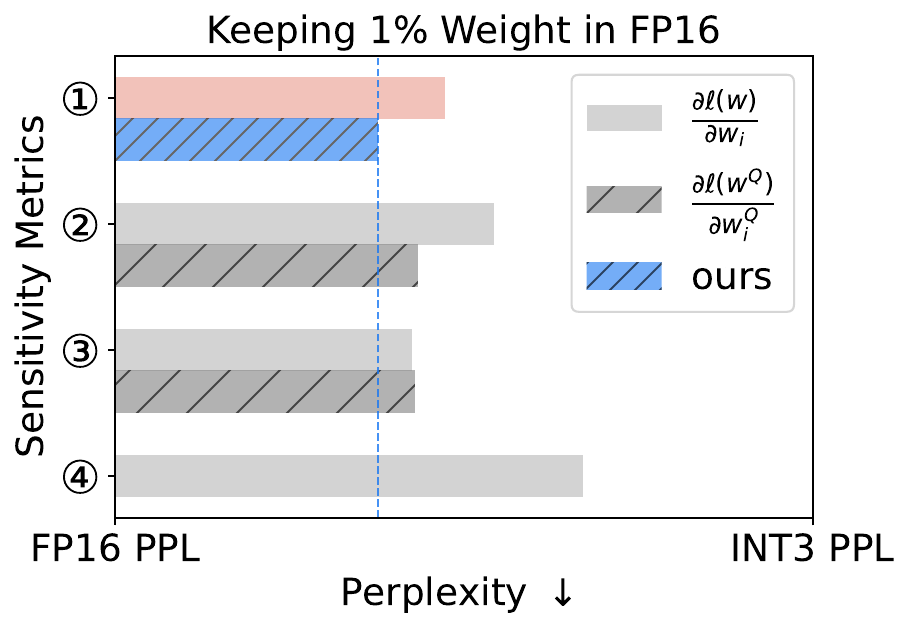}
    \vspace{-.1in}
    \captionof{figure}{The gain (measured in perplexity) of keeping 1\% weights in FP16, selected using different sensitivity metrics.}
    \label{fig:weight_sensitivity}
\end{figure}

In addition, we evaluate our proposed sensitivity metric in fine granularity, where $w_i \in \mathbb{R}$ corresponds to a single weight. Figure~\ref{fig:weight_sensitivity} compares our metric with existing alternatives (as in Table~\ref{tab:sensitivity}), as well as their variants that use gradients computed at quantized weights. We use a fixed mixed-precision protocol that keeps the 1\% most sensitive weights in FP16 while quantizing the rest to INT3, with sensitive weights selected according to different metrics. Among all combinations, first-order gradients obtained around the quantized model consistently outperform the ones obtained at the full-precision model, yielding the largest perplexity gains.

\subsection{Spatial Distribution of Sensitive Weights}\label{app:spatial}
\begin{figure}[h]
    \centering
    \includegraphics[width=0.78\linewidth]{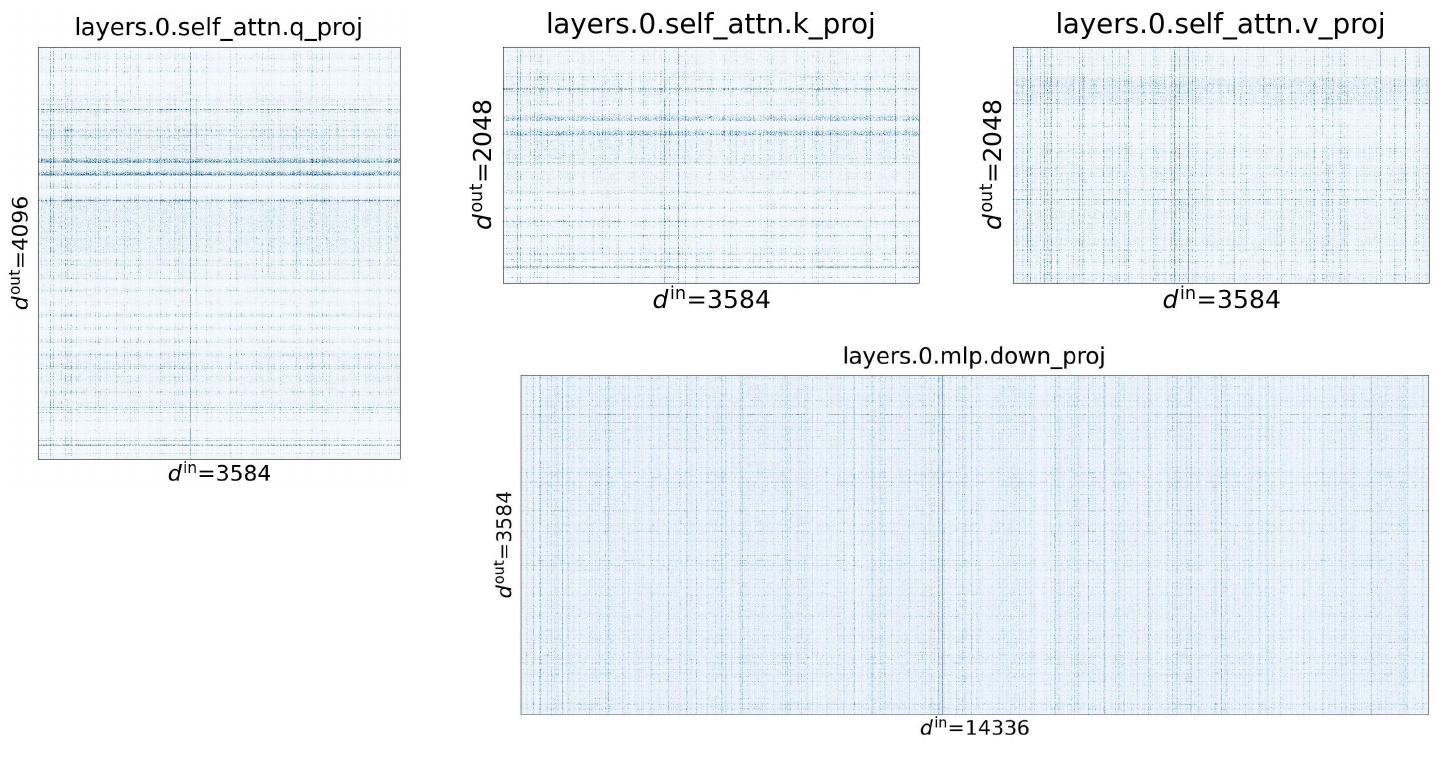} \\
    \includegraphics[width=0.8\linewidth]{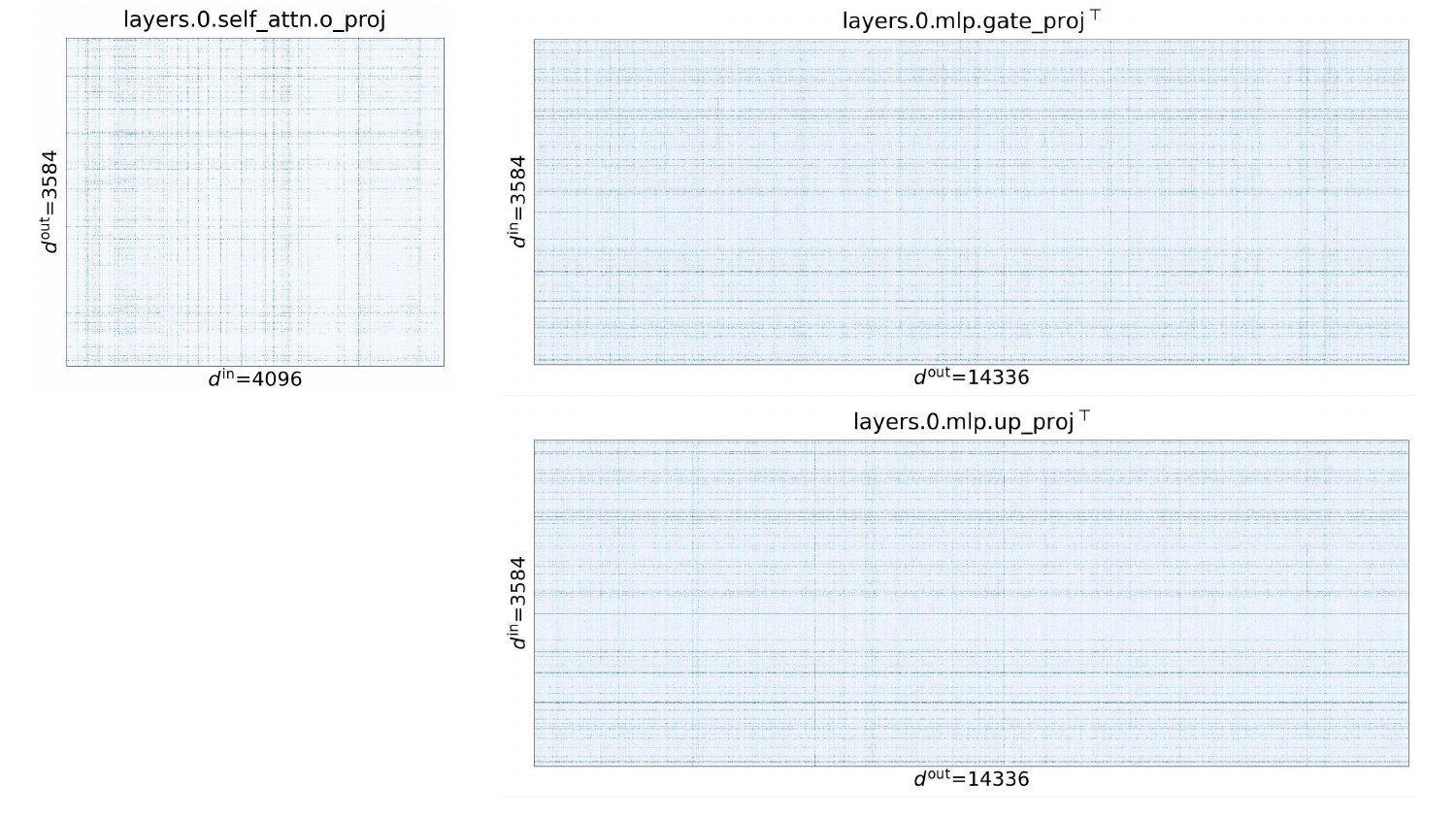}
    \vspace{-.1in}
    \caption{Examples of weight sensitivity distribution in Gemma2-9B.}
    \label{fig:bidirectional}
\end{figure}

Figure~\ref{fig:bidirectional} shows additional examples of weight sensitivity distributions across different layers and models, where bi-directional channel-wise clustering is consistently observed. 

\section{Channel Reordering} \label{app:reorder}
\begin{figure}[h]
    \centering
    \includegraphics[width=0.8\linewidth]{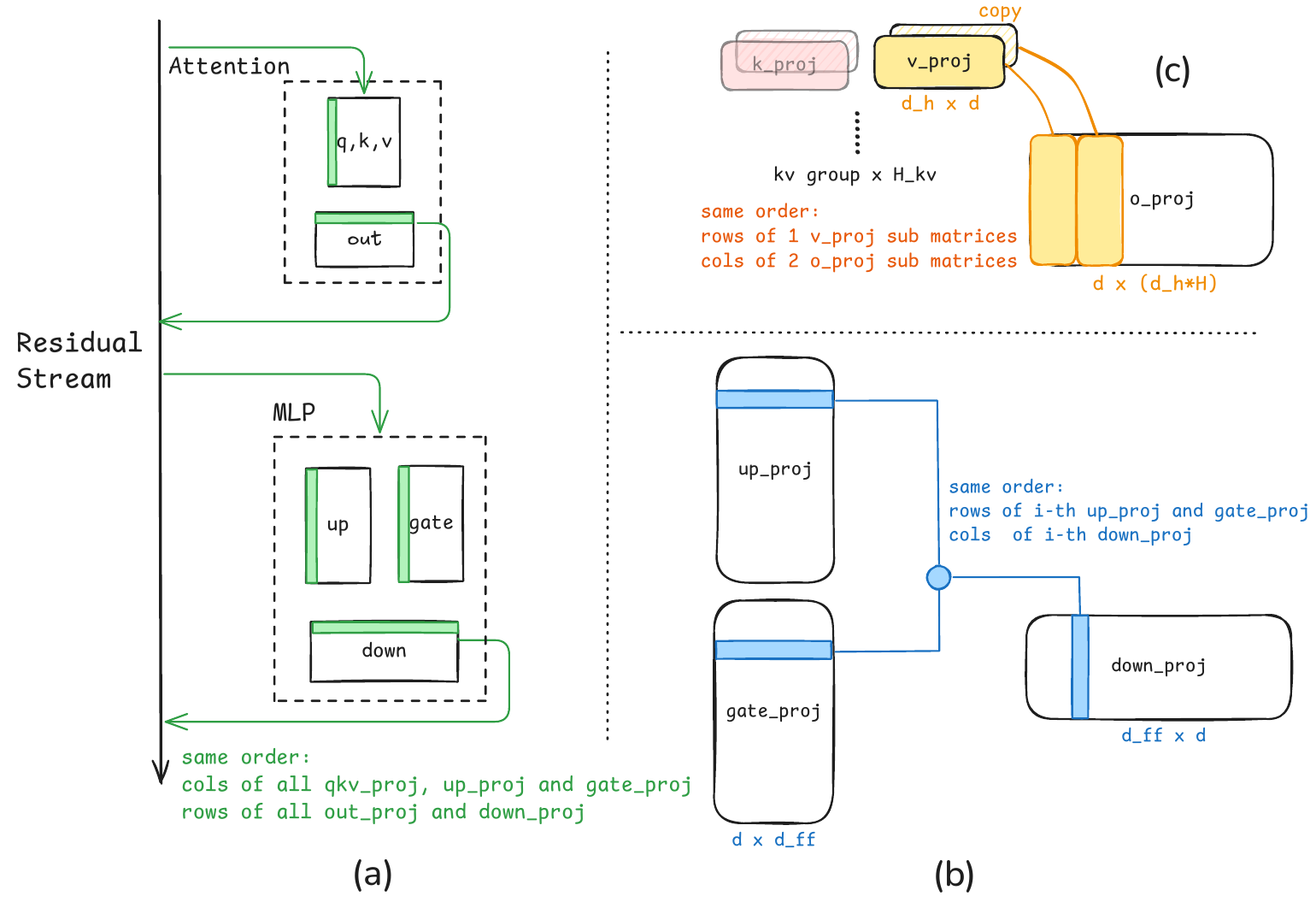}
    \vspace{-.1in}
    \caption{Channel coupling relationships that constrain reordering.}
    \label{fig:reorder_overview}
\end{figure}

As discussed in Section~\ref{sec:partition}, channel reordering can be performed offline as a pre-processing step prior to quantization, as long as the output-input channel ordering is kept consistent across connected layers. The permuted weights can be stored directly without introducing any additional runtime operations. In Transformer architectures, such channel dependencies follow fixed structural patterns, and channels that are structurally coupled tend to exhibit similar sensitivity profiles. Figure~\ref{fig:reorder_overview} illustrates the main types of channel-coupling relationships that constrain reordering. Throughout, weight matrices are represented with shape $d_{\text{out}} \times d_{\text{in}}$, where rows correspond to output channels, and columns correspond to input channels. 

\begin{figure}
    \centering
    \includegraphics[width=0.98\linewidth]{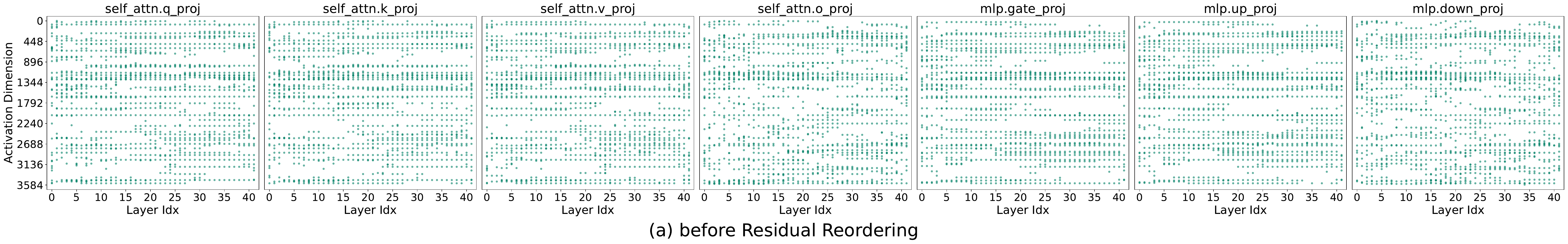}
    \includegraphics[width=0.98\linewidth]{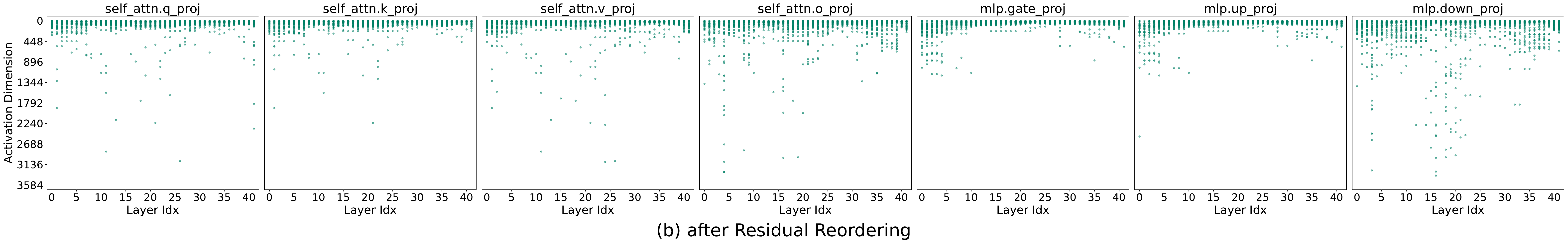}
    \caption{The top 1\% sensitive channels along the residual stream (a) before and (b) after joint reordering.}
    \label{fig:reorder_global}
\end{figure}

We first consider the global coupling induced by the residual stream, which corresponds to the shared hidden activation dimensions propagated across all Transformer layers. As shown in Figure~\ref{fig:reorder_overview}~(a), any layer that consumes or produces the residual stream must preserve a consistent index ordering of these hidden dimensions to maintain functional equivalence. In addition, we shuffle and store the weights of the embedding head accordingly, to preserve model input and output.
Figure~\ref{fig:reorder_global} visualizes the locations of the top 1\% most sensitive channels across all weight matrices connected to the residual stream. Prior to reordering, these sensitive channels are dispersed across indices but remain aligned across layers and projections. After applying a joint reordering—computed by aggregating sensitivity scores across all coupled matrices—the sensitive channels become tightly clustered, revealing a coherent global structure shared along the residual pathway.

Beyond the residual stream, channel couplings within attention and MLP blocks (related to intermediate activations) are more localized and impose weaker reordering constraints. In MLP layers (see Figure~\ref{fig:reorder_overview}~(b)), reordering only needs to be consistent between three projections (e.g., up, gate, and down), allowing different MLP blocks to be reordered independently.

In multi-head attention (see Figure~\ref{fig:reorder_overview}~(c)), local channels are partitioned into groups corresponding to attention heads, and reordering must preserve the internal structure of each head. As a result, reordering is restricted to operate independently within each head group. In practice, we apply local reordering only to the value and output projections, while keeping the output channel order of query and key projections, because of the complex constraints by positional encoding mechanisms such as RoPE \cite{su2024roformer} and by shared normalization \textit{qk\_norm} (e.g., in Qwen3 \cite{yang2025qwen3}). Since the QK projections account for only a small fraction of the total parameters, excluding their local reordering has negligible impact on overall effectiveness.


\begin{figure}[h]
    \centering
    \includegraphics[width=0.51\linewidth]{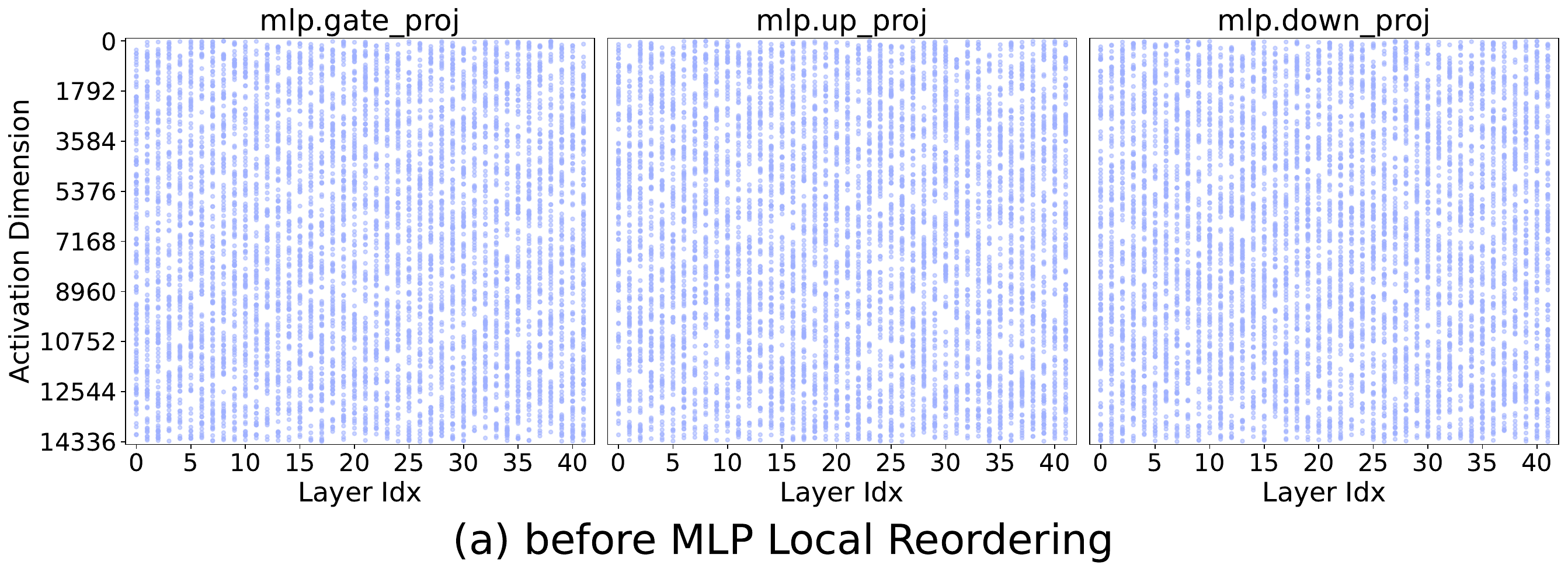}
    \includegraphics[width=0.34\linewidth]{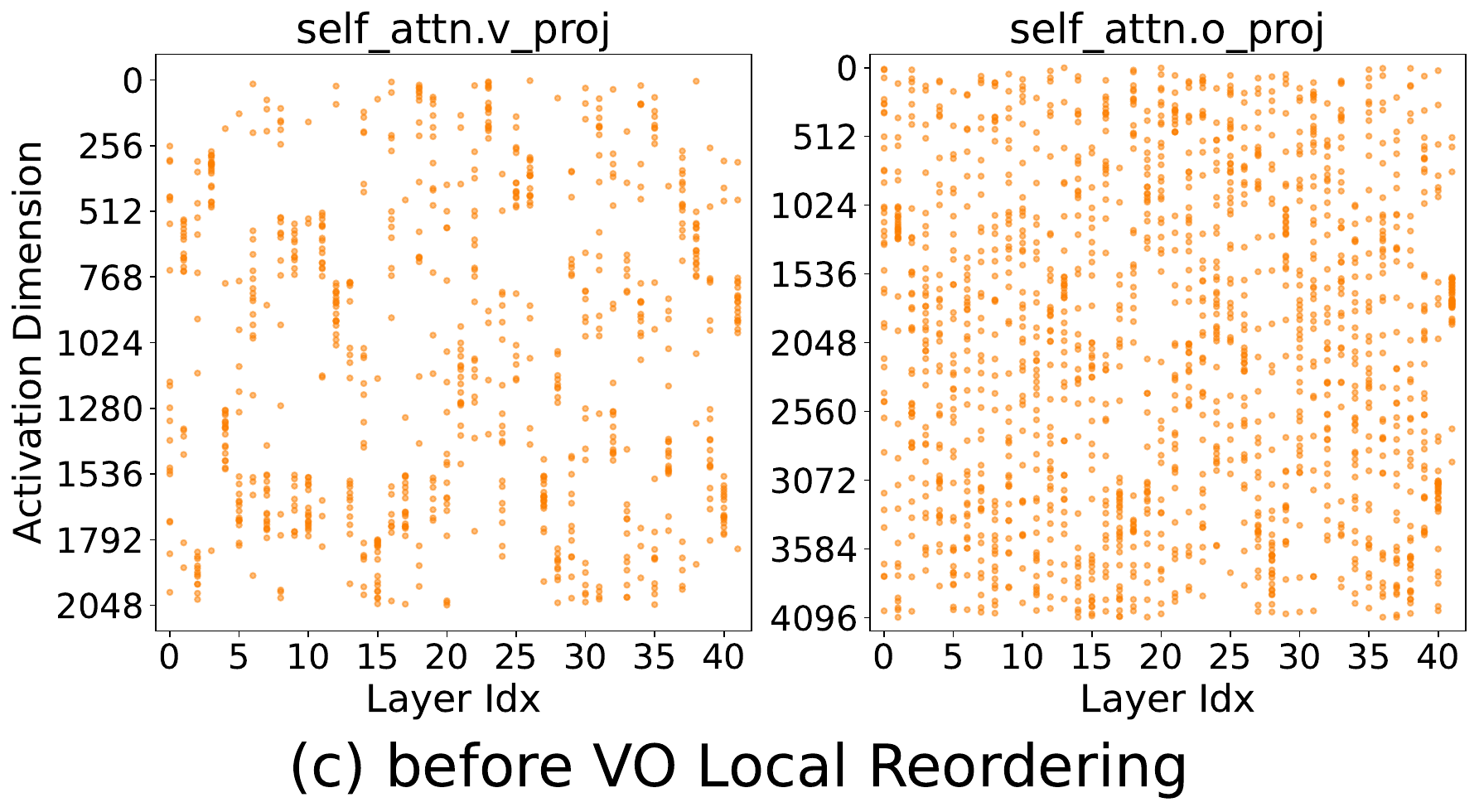} \\
    \includegraphics[width=0.51\linewidth]{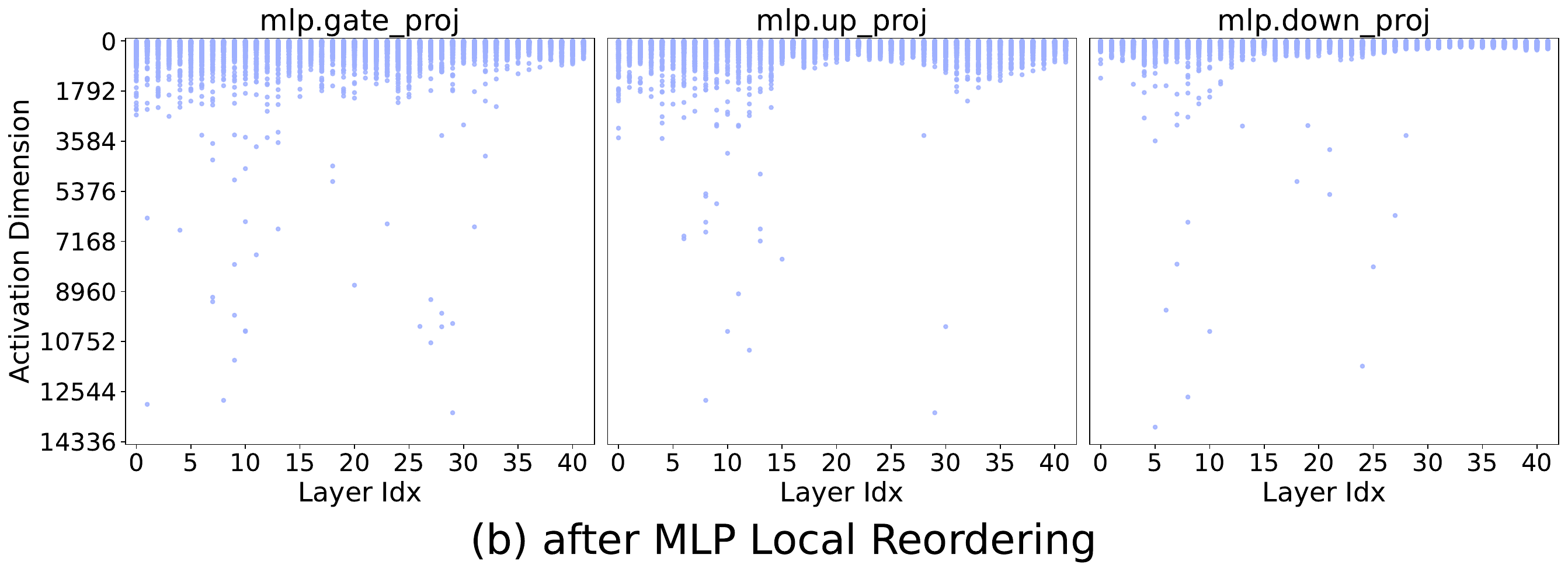}
    \includegraphics[width=0.34\linewidth]{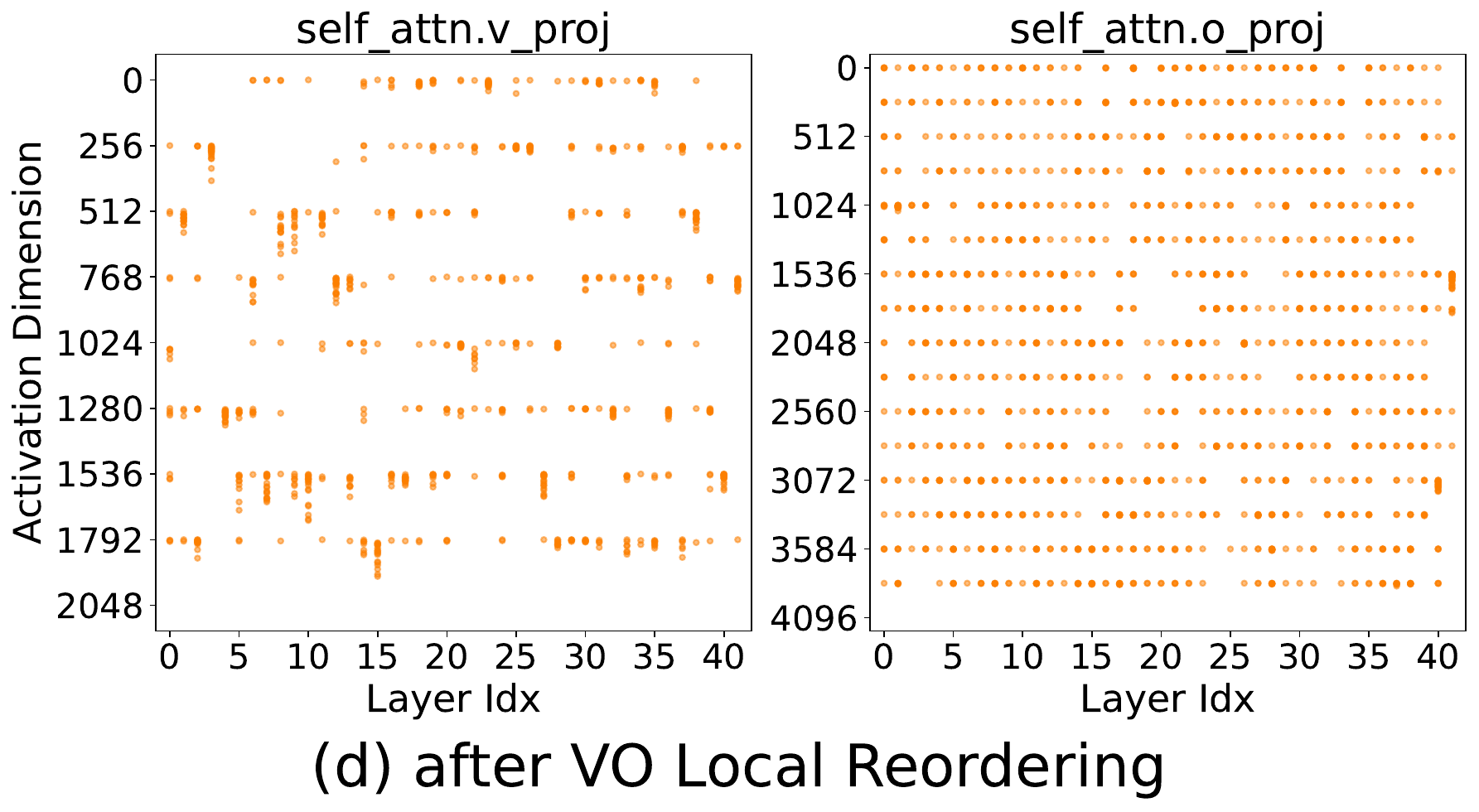}
    \caption{The top 1\% sensitive channels inside MLP or attention blocks (a)(c) before and (b)(d) after local reordering. Note that for attention projections, sensitive channels are clustered within each attention head (instead of the whole matrix) as expected.}
    \label{fig:reorder_local}
\end{figure}

\newpage
\section{Ablation Study}\label{app:ablation}
In this section, we present the ablation study of \ours using the Gemma2-9B model in the 3-4 bit regime.

\subsection{impact of gradient update}
\begin{figure}[h]
    \vspace{.05in}
    \centering
    \includegraphics[width=0.32\linewidth]{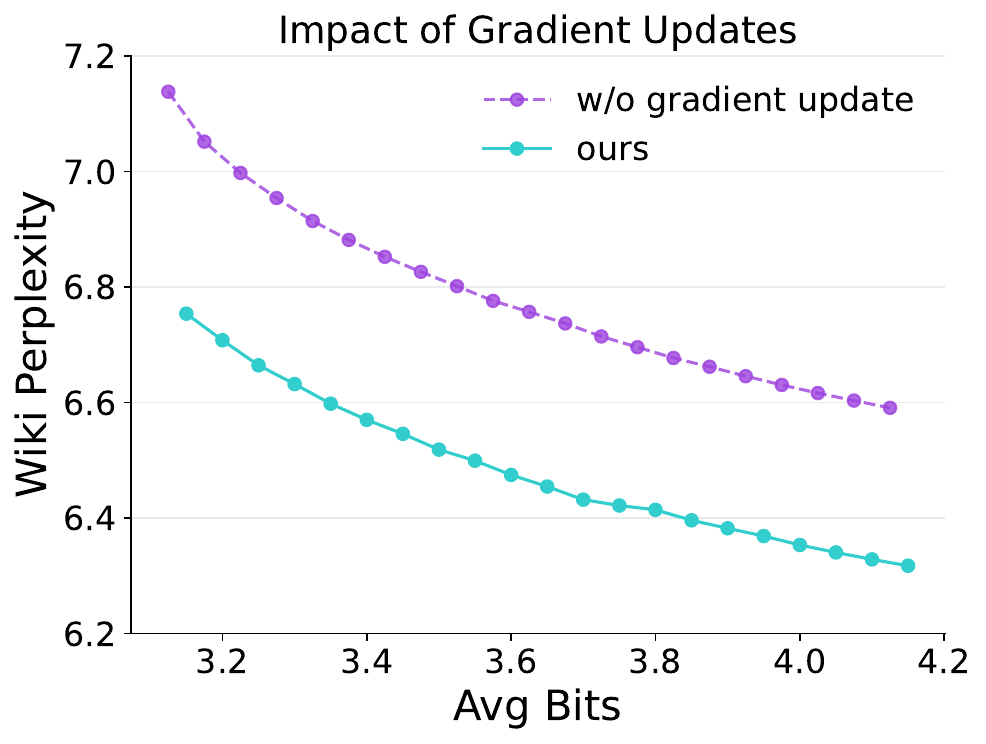}
    \hspace{0.05\linewidth}
    \includegraphics[width=0.32\linewidth]{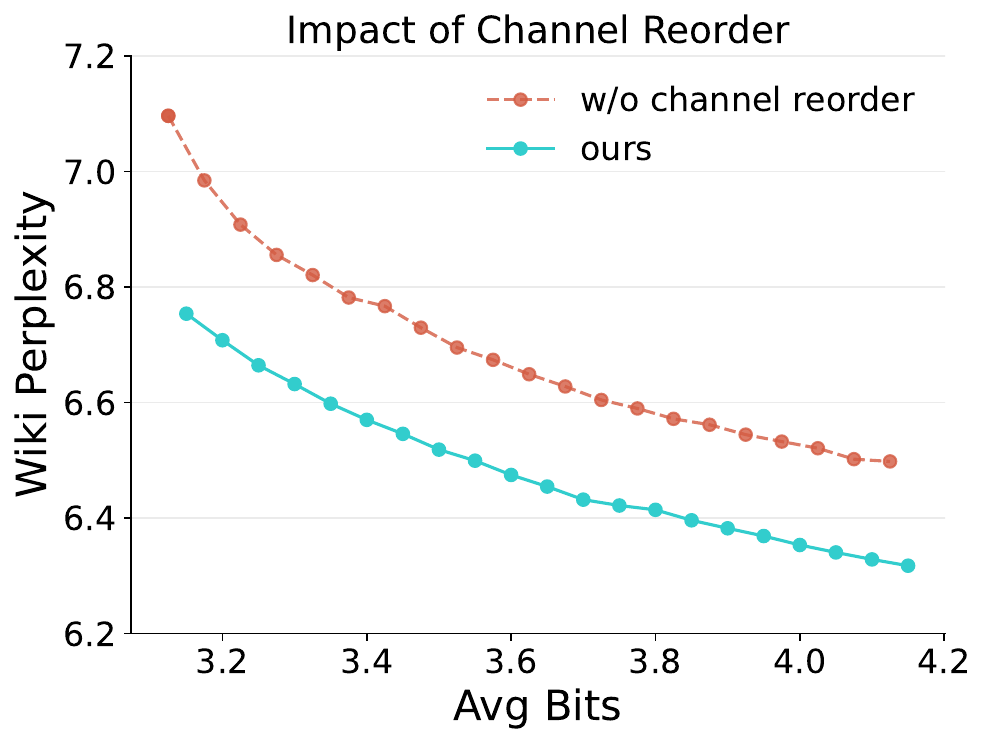}
    \caption{Ablation study on the impact of adaptive gradient updates and channel reordering.}
    \label{fig:grad_reorder}
\end{figure}

As discussed in Section~\ref{sec:sensitivity}, weight sensitivity is inherently dynamic. In Algorithm~\ref{alg:sgreedy}, we therefore update the sensitivity estimates at each iteration using gradients computed from the current quantized model. To ablate the effect of this adaptive sensitivity estimation, we consider a variant that fixes the gradients obtained at the first iteration. As shown in Figure~\ref{fig:grad_reorder} (left), adaptive updating consistently yields better performance.

\subsection{impact of channel reordering}
Figure~\ref{fig:grad_reorder} (right) further verifies the effectiveness of channel reordering. A clear performance gap is observed when reordering is disabled, demonstrating the importance of aligning weight layout with sensitivity structure.

\subsection{choice of sensitivity statistics}
Algorithm~\ref{alg:sgreedy} requires an efficient estimate of the marginal loss change when increasing or decreasing the precision of a weight block. Ideally, using the first-order expansion around the current quantized model $\vw^{Q_b}$ (see Section~\ref{sec:sensitivity}), the loss changes can be approximated as ${\vg^{(\vw^{Q_b})}}^\top (\vw^{Q_{b\pm1}}-\vw^{Q_{b}})$. However, computing the exact perturbation from a one-bit change requires additional re-quantization of the block. To avoid this overhead, we adapt two practical surrogates.

Increasing precision primarily reduces existing quantization error. The correction introduced by a finer quantizer is bounded by the current error and, in expectation, moves the quantized weight toward the full-precision one $\vw$. We therefore approximate the gain of increasing precision for block $i$ as 
\begin{equation}
    (s_{\text{up}})_i = {\vg^{(\vw_i^{Q_{b_i}})}}^\top (\vw_i -\vw_i^{Q_{b_i}}).
\end{equation}

Decreasing precision introduces additional distortion due to the larger quantization step size. Since the direction of this perturbation is unpredictable and its magnitude scales with the quantization step size $\Delta_{b_i} \propto 2^{-b_i}$, we instead estimate the expected magnitude of the loss increase using
\begin{equation}
    (s_{\text{down}})_i = \epsilon_{b_i}\|\vg^{(\vw_i^{Q_{b_i}})} \odot \vw_i^{Q_{b_i}}\|_1, \quad\quad \epsilon_{b_i} = \frac{1}{2^{b_i}}.
\end{equation}

\begin{figure}[h]
    \centering
    \includegraphics[width=0.36\linewidth]{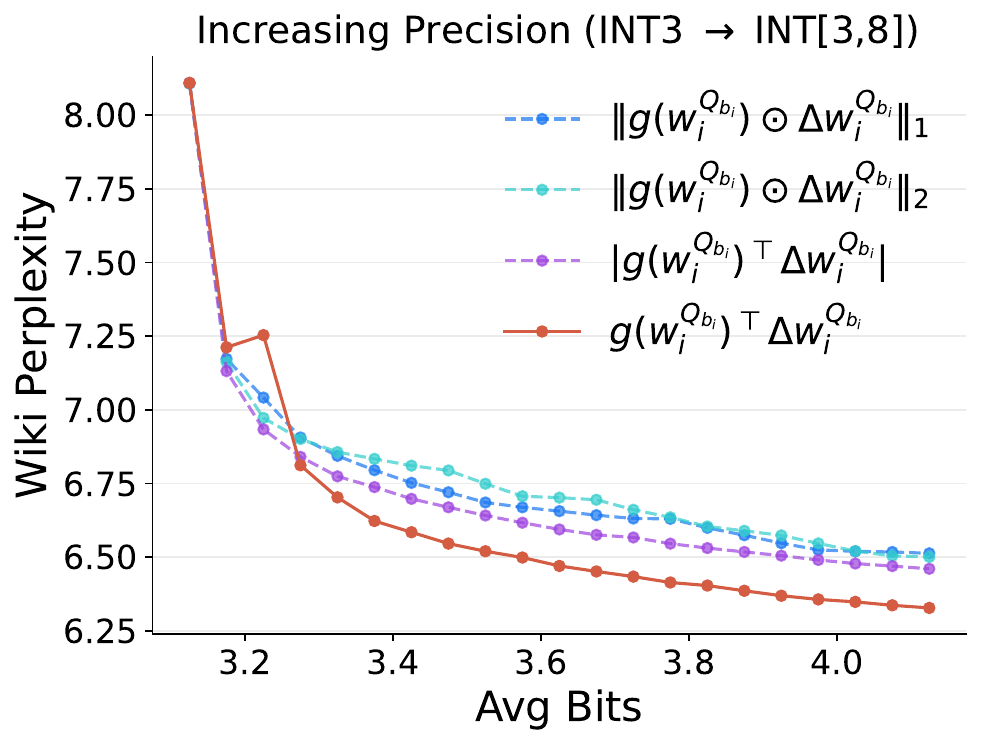}
    \hspace{0.03\linewidth}
    \includegraphics[width=0.5\linewidth]{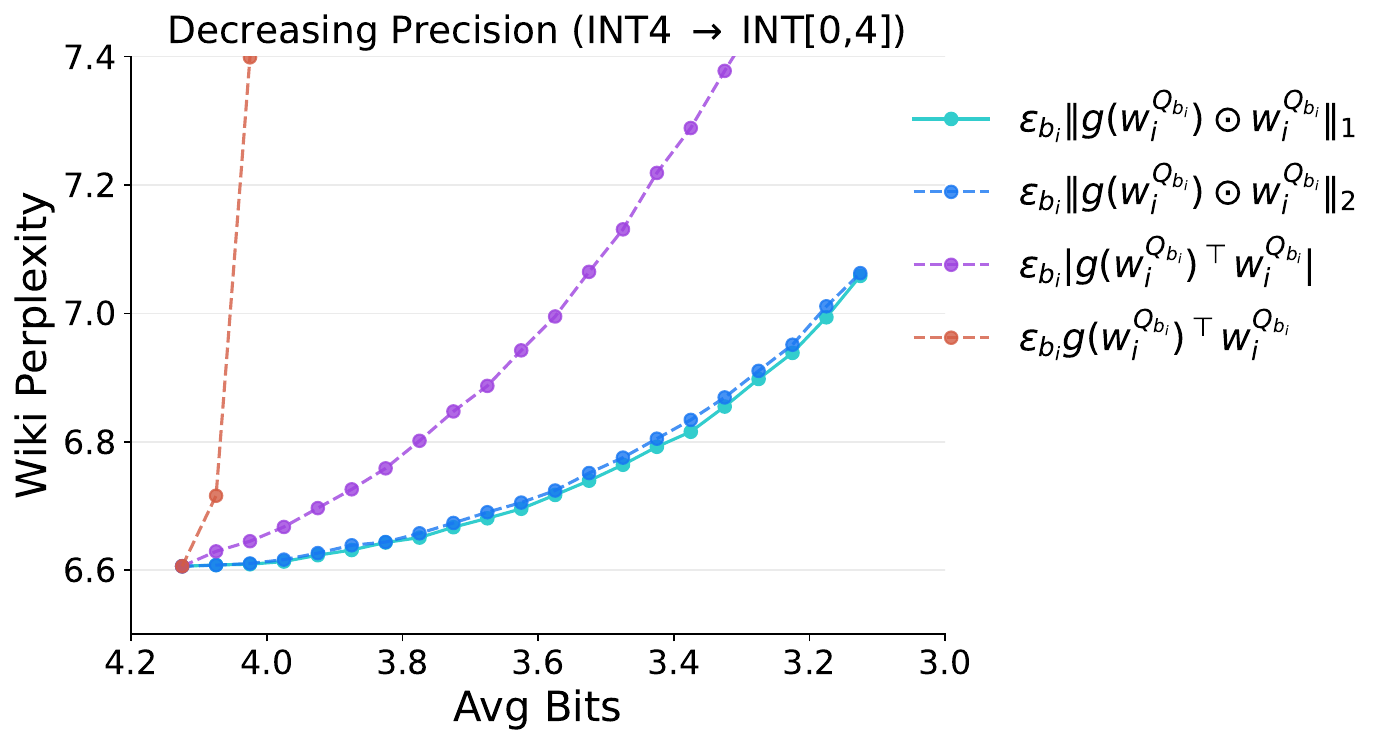}
    \caption{Ablation on sensitivity statistics for one-sided precision updates.}
    \label{fig:sensitivity_stats}
\end{figure}

Figure~\ref{fig:sensitivity_stats} empirically supports this asymmetric modeling. We evaluate different sensitivity statistics by performing the greedy updates using only precision increases (left) or only precision decreases (right). Signed aggregation works best when increasing precision, while magnitude-based measures ($l1$ and $l2$ norms) better capture degradation when decreasing precision.

% \subsection{[warm start] impact of initialization}

\subsection{impact of batch size $k$}
\begin{figure}[h]
    \centering
    \includegraphics[width=0.31\linewidth]{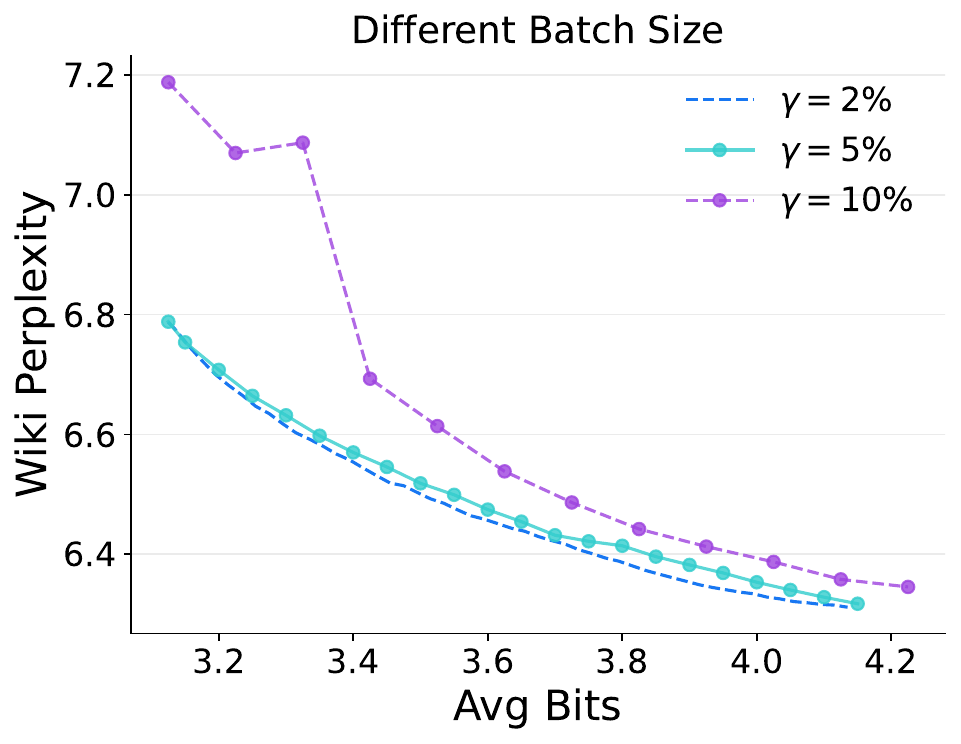}
    \hspace{0.01\linewidth}
    \includegraphics[width=0.31\linewidth]{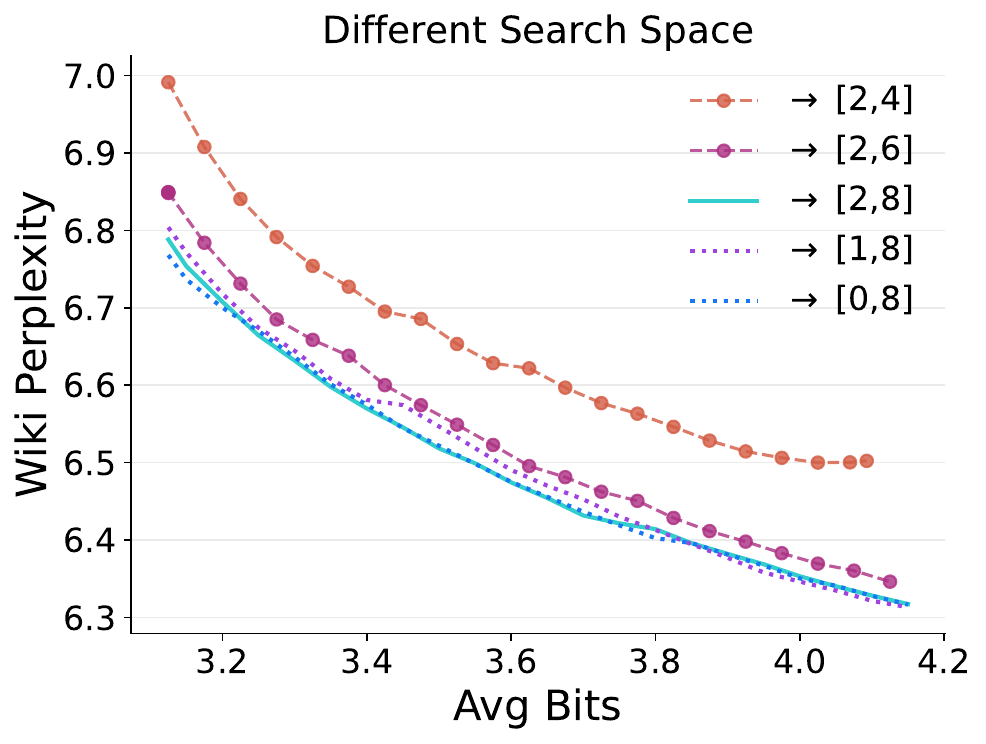}
    \hspace{0.01\linewidth}
    \includegraphics[width=0.31\linewidth]{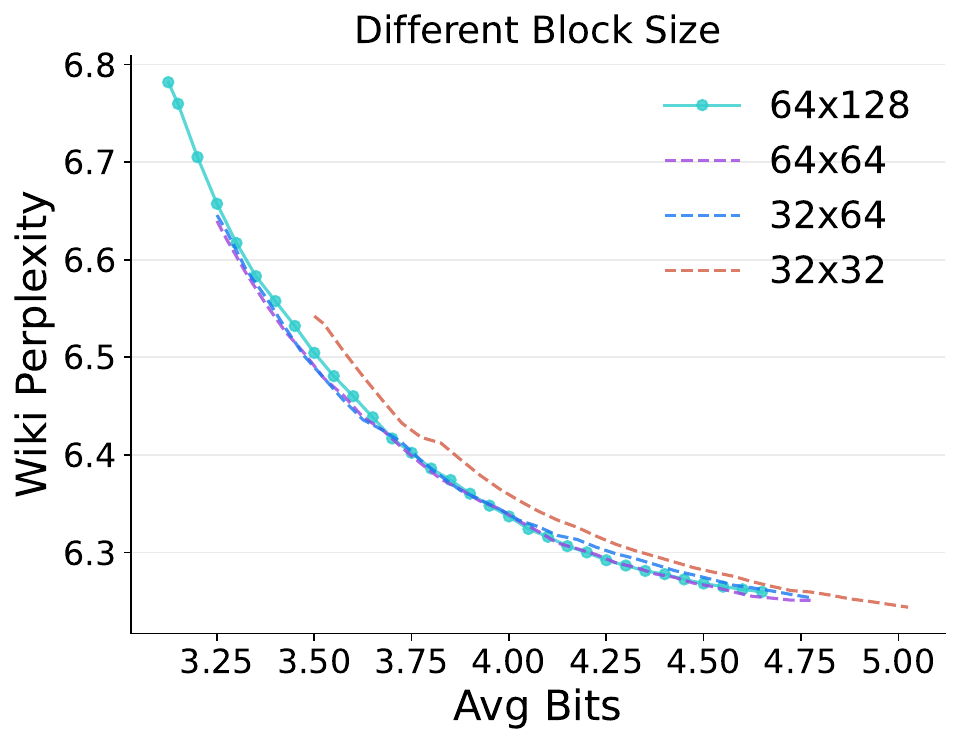}
    \caption{Ablation study on the choices of batch size (left), precision search space (middle) and weight block size (right).}
    \label{fig:batch_space}
\end{figure}

The batch size $k=\lfloor \gamma N\rfloor$ in Algorithm~\ref{alg:sgreedy} controls the trade-off between convergence speed and search quality. Figure~\ref{fig:batch_space} (left) compares different choices of $\gamma$. We observe that a large batch size (e.g., $\gamma=10\%$) leads to noticeable performance degradation, while small values such as $5\%$ and $2\%$ yield similar perplexity at the same bitwidth. To balance efficiency and quality, we set the initial update rate $\gamma_0=5\%$ in all main experiments.

\subsection{impact of precision search space}
Figure~\ref{fig:batch_space} (middle) examines how restricting the precision search space affects performance. We compare configurations with different minimum and maximum bitwidths. Limiting the maximum precision to 4 bits consistently degrades performance, indicating that a small subset of highly sensitive weights benefits from higher precision. In contrast, allowing lower precisions down to 0 or 1 bit has minimal impact, as the resulting tradeoff curves largely overlap. This suggests that there exist sufficient low-sensitivity weights that can be aggressively quantized to offset the higher precision allocation to sensitive components. Overall, flexibility in assigning higher precision to critical weights matters more than expanding the search space to the low-bit end. 

\subsection{impact of block size}
Recall that when partitioning weights into blocks, the block size must align with the tile size of the underlying matrix multiplication kernel. In our experiments, we use $64 \times 128$ based on our Triton kernel implementation. Figure~\ref{fig:batch_space} (right compares several commonly used tile sizes and shows that they yield similar accuracy-compression tradeoffs. Note that the quantization group size must match the block width (number of columns). When the block size is small, more groups are required, increasing the storage overhead of group-wise quantization parameters. As a result, smaller block sizes do not improve accuracy at the same effective bitwidth.

\section{Other Visualization}\label{app:visual}
\begin{figure}[h]
    \vspace{.05in}
    \centering
    \includegraphics[width=0.5\linewidth]{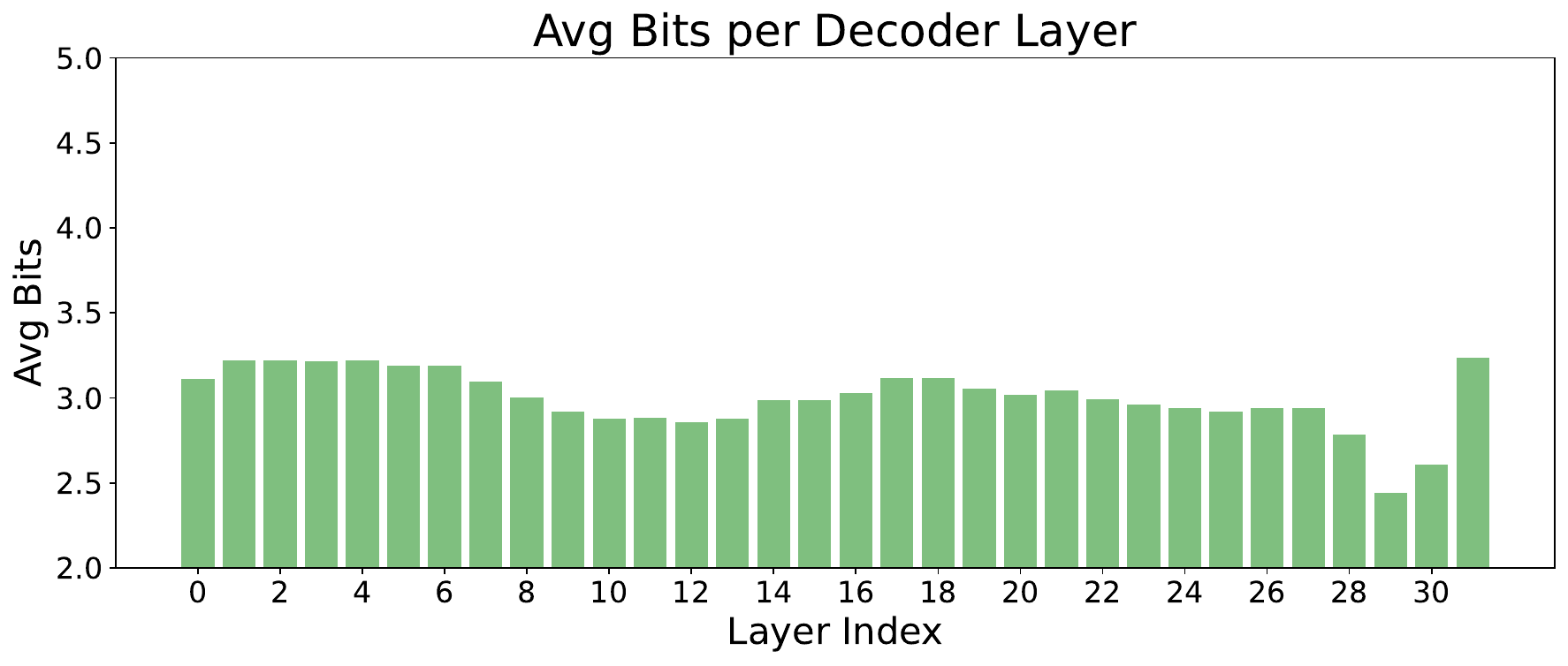} \\
    \includegraphics[width=0.8\linewidth]{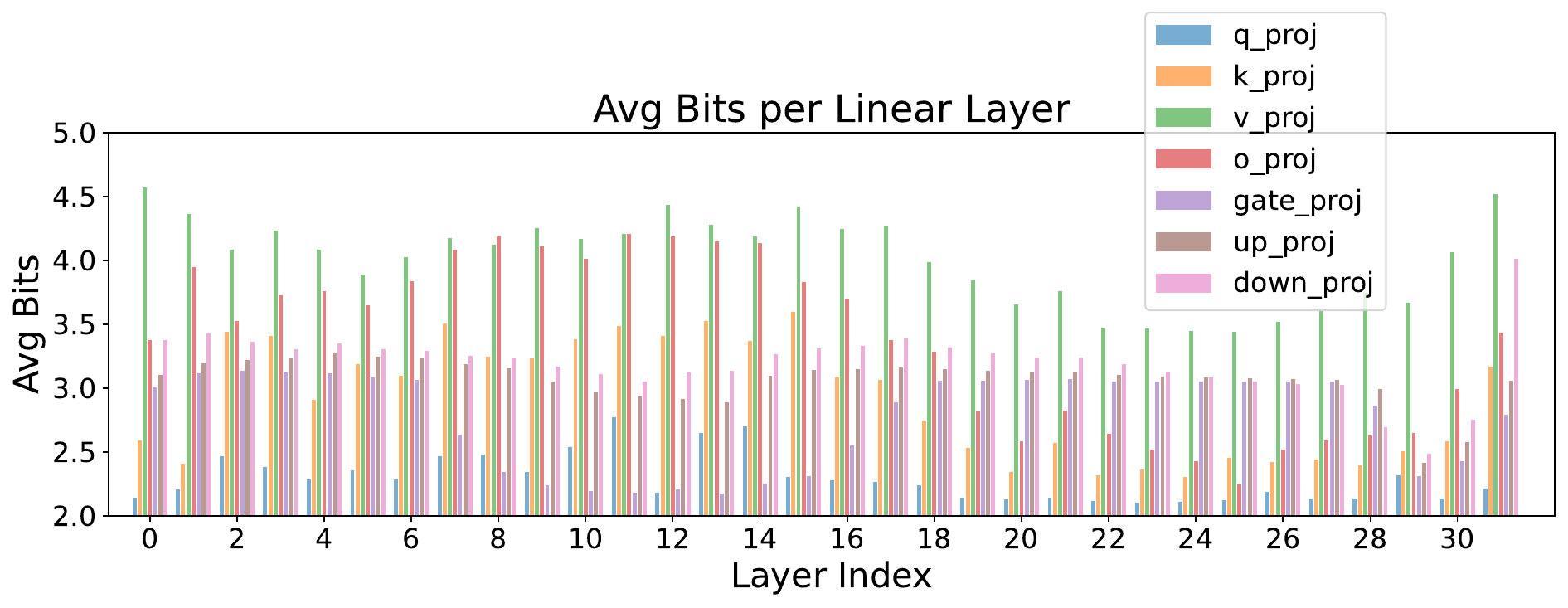}
    \caption{The average number of bits assigned to each decoder (up) and linear layer (down) of Llama3-8B after applying \ours, with a global budget of 3 bits per weight.}
    \label{fig:linear_bit}
\end{figure}

We provide additional analysis of the mixed-precision allocation learned by \ours using the same 3-bit Llama3-8B case study presented in Section~\ref{sec:ablation}. Beyond the sensitivity redistribution and block-level patterns shown in Figures~\ref{fig:itr_sensitivity} and~\ref{fig:block_precision}, we further examine how precision is distributed across layers in the final allocation. Figure~\ref{fig:linear_bit} (top) shows the average bitwidth assigned to each decoder layer. Although the first and last layers exhibit dominant sensitivity under uniform-precision quantization, the learned allocation does not simply assign substantially higher average precision to these layers. Instead, the per-layer averages remain relatively smooth, indicating that precision reallocation occurs primarily within layers rather than through coarse layer-wise trade-offs.

Zooming in to individual linear projections (Figure~\ref{fig:linear_bit}, bottom), we observe clear heterogeneity across layer types. In particular, \texttt{v\_proj} layers in attention consistently receive higher average precision across depth, suggesting that they are more sensitive to quantization errors. In contrast, \texttt{q\_proj} tends to receive lower precision.

\section{Other Experiment Results}\label{app:results}
Table~\ref{tab:other_mp} compares \ours with other mixed-precision approaches. Since several earlier methods report results only on Llama2 models with context length 2048. We here evaluate \ours under the same setting for fair comparison. Under comparable effective bitwidths, \ours consistently outperforms existing baselines. 

The compared methods apply mixed-precisionat different granularities, and element-wise schemes typically introduce additional index storage cost and runtime overhead. In particular, AMQ automates layer-wise precision search. PB-LLM and SqueezeLLM store sensitive weights in high precision. SlimLLM+ extends SlimLLM by finetuning the quantization parameters using the OmniQuant recipe. ICQuant does not perform explicit mixed-precision allocation; instead, it applies a separate codebook to outliers (with large magnitude), where additional memory overhead arises from storing their indices.

We next evaluate the generalization of our mixed-precision framework. Table~\ref{tab:L3-it} extends the evaluation to an instruction-tuned model on more challenging, reasoning-intensive benchmarks. Table~\ref{tab:gemma} examines a Gemma-2 model, which exhibits distinct sensitivity characteristics compared with Llama models, as illustrated in Figure~\ref{fig:layer_sensitivity}. In these experiments, we compare \ours against its uniform-precision RTN backend to isolate the benefit of precision allocation. In all settings, \ours consistently outperforms its uniform-precision counterpart, with up to 42\% increase in Gsm8k accuracy.

\begin{table}[h]
    \vspace{.1in}
    \centering
    \resizebox{0.58\textwidth}{!}{
    \begin{tabular}{c|c|c|c|c|c}
        \toprule
        \multirow{2}{*}{Method} & \multirow{2}{*}{MP} & \multirow{2}{*}{Granularity} & \multirow{2}{*}{bits} & \multicolumn{2}{c}{Llama2 - 7B} \\
         & & & & Wiki2 $\downarrow$ & C4 $\downarrow$ \\
         \midrule 
         AMQ & $\checkmark$ & layer & 2.5 & 9.24 & 12.37 \\
         PB-LLM & $\checkmark$ & element & 2.5 & 24.53 & 32.05 \\
         SqueezeLLM ($\star$) & $\checkmark$ & element & 2.2 & 10.79 & | \\
         SlimLLM+ & $\checkmark$ &channel group & 2.1 & 10.87 & 18.18 \\
         \myrowcolour% 
         \ours + RTN & $\checkmark$ &block & 2.1 & \textbf{8.08} & \textbf{9.81} \\
         \midrule
         ICQuant ($\star$) & $\circ$ & element & 2.3 & 7.21 & 8.97 \\
         \myrowcolour% 
         \ours + RTN & $\checkmark$ & block & 2.3 &  \textbf{7.15} & \textbf{8.84} \\
         \bottomrule
    \end{tabular}
    }
    \caption{Evaluation of mixed-precision Llama2-7B model in 2-2.5 bit regime. Wiki2 and C4 denotes perplexity on WikiText-2 and C4 with context length 2048. Methods marked with $\star$ use non-uniform quantizer; all others use uniform quantizers. MP indicates mixed-precision.}
    \label{tab:other_mp}
\end{table}

\begin{table}[h]
    \centering
    \resizebox{0.7\textwidth}{!}{
    \begin{tabular}{c|c|c|c|c|c|c|c}
        \toprule
        \multirow{2}{*}{Method} & \multirow{2}{*}{MP} & \multirow{2}{*}{bits} & \multicolumn{5}{c}{Llama3.1 8B-Instruct} \\
         & & & Wiki2 $\downarrow$ & Zero-shot & MMLU & Gsm8k & MBPP\\
         \midrule
         - & $\times$ & 16 & 6.50 & 71.85 & 68.25 & 85.52 & 58.00 \\ 
         \midrule
         RTN-g128& $\times$ & 3.1 & 11.13 & 63.70 & 52.14 & 29.04 & 30.80 \\
         % GPTQ-g128 & $\times$ & 3.1 &  &  \\
         \myrowcolour% 
         \ours + RTN & $\checkmark$ & 3.1 & 7.32 & 69.23 &  63.30 & 72.18 & 44.60 \\
         \midrule
         RTN-g128& $\times$ & 2.1 & 1e6 & 37.45 & 24.78 & 0 & 0\\
         % GPTQ-g128 & $\times$ & 2.1 &   \\
         \myrowcolour% 
         \ours + RTN & $\checkmark$ & 2.1 & 12.75 & 61.79 & 50.38 & 20.09 & 14.80 \\
         \bottomrule
    \end{tabular}
    }
    \caption{Evaluation of instruction-tuned Llama3.1-8B under 2-3 bit regime. Wiki2 reports perplexity on WikiText-2 (context length = 8192). 0-shot denotes average accuracy over six zero-shot tasks. MP indicates mixed-precision.}
    \label{tab:L3-it}
\end{table}

\begin{table}[h]
    \centering
    \resizebox{0.55\textwidth}{!}{
    \begin{tabular}{c|c|c|c|c|c}
        \toprule
        \multirow{2}{*}{Method} & \multirow{2}{*}{MP} & \multirow{2}{*}{bits} & \multicolumn{3}{c}{Gemma2 - 9B}  \\
         & & & Wiki2 $\downarrow$ & Zero-shot & MMLU \\
         \midrule
         - & $\times$ & 16 & 6.36 & 74.89 & 70.50 \\ 
         \midrule
         RTN-g128& $\times$ & 3.1 & 8.11 & 71.40 &  65.52 \\
         % GPTQ-g128 & $\times$ & 3.1 &  &  \\
         \myrowcolour% 
         \ours + RTN & $\checkmark$ & 3.1 & 6.74 & 72.86 & 67.12\\
         \midrule
         RTN-g128& $\times$ & 2.1 & 3e4 & 39.31 & 23.96 \\
         % GPTQ-g128 & $\times$ & 2.1 &   \\
         \myrowcolour% 
         \ours + RTN & $\checkmark$ & 2.1 & 8.89 & 67.05 & 57.89 \\
         \bottomrule
    \end{tabular}
    }
    \caption{Evaluation of quantized Gemma2-9B under 2-3 bit regime. Wiki2 reports perplexity on WikiText-2 (context length = 4096). 0-shot denotes average accuracy over six zero-shot tasks. MP indicates mixed-precision.}
    \label{tab:gemma}
\end{table}

%%%%%%%%%%%%%%%%%%%%%%%%%%%%%%%%%%%%%%%%%%%%%%%%%%%%%%%%%%%%%%%%%%%%%%%%%%%%%%%
%%%%%%%%%%%%%%%%%%%%%%%%%%%%%%%%%%%%%%%%%%%%%%%%%%%%%%%%%%%%%%%%%%%%%%%%%%%%%%%

\end{document}

%% file: math_commands.tex
\usepackage{amsmath}
\usepackage{amssymb}
\usepackage{amsfonts,bm}
 \usepackage{bbm}

\usepackage{hyperref}
 
\usepackage{xcolor}
\usepackage{soul}
\usepackage{thmtools, thm-restate}
\newcounter{remarkcnt}

\newcounter{interpcnt}

\def\eqref#1{equation~\ref{#1}}
\def\floor#1{\lfloor #1 \rfloor}
\def\1{\bm{1}}

\def\vzero{{\bm{0}}}

\def\vb{{\bm{b}}}

\def\ve{{\bm{e}}}

\def\vg{{\bm{g}}}

\def\vs{{\bm{s}}}

\def\vw{{\bm{w}}}
\def\vx{{\bm{x}}}
\def\vy{{\bm{y}}}

\def\mW{{\bm{W}}}
\def\mX{{\bm{X}}}

\DeclareMathAlphabet{\mathsfit}{\encodingdefault}{\sfdefault}{m}{sl}
\SetMathAlphabet{\mathsfit}{bold}{\encodingdefault}{\sfdefault}{bx}{n}

\def\gB{{\mathcal{B}}}

\def\gD{{\mathcal{D}}}

\def\gF{{\mathcal{F}}}